\pdfoutput=1

\documentclass[11pt]{article}

\usepackage[final]{acl}

\usepackage{times}
\usepackage{latexsym}
\usepackage{multirow}

\usepackage[T1]{fontenc}
\usepackage{colortbl}

\usepackage[utf8]{inputenc}

\usepackage{microtype}
\usepackage{hyperref}
\usepackage{graphicx}
\usepackage{amsmath, amsthm, amssymb}
\usepackage{bbm}
\usepackage{booktabs}
\usepackage{caption}
\usepackage{url}
\usepackage{enumerate, enumitem}
\usepackage{newtxtext, newtxmath}
\usepackage{subfig}
\usepackage[most]{tcolorbox}
\usepackage{array}
\usepackage{tabularx}

\newcommand\blfootnote[1]{%
  \begingroup
  \renewcommand\thefootnote{}\footnote{#1}%
  \addtocounter{footnote}{-1}%
  \endgroup
}

\title{Evaluating Evaluation Metrics -- The \textit{Mirage} of Hallucination Detection}

\author{
 \textbf{Atharva Kulkarni\textsuperscript{$\spadesuit\ast$}}
 \quad\textbf{Yuan Zhang\textsuperscript{$\diamondsuit$}}
 \quad\textbf{Joel Ruben Antony Moniz\textsuperscript{$\diamondsuit$}}
 \quad\textbf{Xiou Ge\textsuperscript{$\diamondsuit$}}
 \\ 
 \textbf{Bo-Hsiang Tseng\textsuperscript{$\diamondsuit$}}
 \quad\textbf{Dhivya Piraviperumal\textsuperscript{$\diamondsuit$}}
\quad\textbf{Swabha Swayamdipta\textsuperscript{$\spadesuit$}}
 \quad\textbf{Hong Yu\textsuperscript{$\diamondsuit$}}
 \\
\textsuperscript{$\spadesuit$}University of Southern California \quad
\textsuperscript{$\diamondsuit$}Apple Inc.
\\
\textbf{Correspondence:} \href{mailto:atharva.kulkarni@usc.edu}{atharva.kulkarni@usc.edu}, \href{mailto:yzhang73@apple.com}{yzhang73@apple.com}
}

\begin{document}
\maketitle

\begin{abstract}
Hallucinations pose a significant obstacle to the reliability and widespread adoption of language models, yet their accurate measurement remains a persistent challenge.
While many task- and domain-specific metrics have been proposed to assess \textit{faithfulness} and \textit{factuality} concerns, the robustness and generalization of these metrics are still untested.
In this paper, we conduct a large-scale empirical evaluation of 6 diverse sets of hallucination detection metrics across 4 datasets, 37 language models from 5 families, and 5 decoding methods.
Our extensive investigation reveals concerning gaps in current hallucination evaluation: metrics often fail to align with human judgments, take an overtly myopic view of the problem, and show inconsistent gains with parameter scaling.
Encouragingly, LLM-based evaluation, particularly with GPT-4, yields the best overall results, and mode-seeking decoding methods seem to reduce hallucinations, especially in knowledge-grounded settings.
These findings underscore the need for more robust metrics to understand and quantify hallucinations, and better strategies to mitigate them. 
\end{abstract}
\section{Introduction}
\blfootnote{$^\ast$Work done during internship at Apple Inc.}
Hallucinations in language model generations are detrimental and, unfortunately, a pervasive phenomenon \cite{ji2023survey, kaddour2023challenges, xu2024hallucination}. 
As language models are rapidly adopted across various settings, addressing hallucinations has become a key research focus \cite{varshney2023stitch, dhuliawala2023chain, chuang2024dola, shi-etal-2024-trusting}. 
However, before investing time and resources into devising its mitigation techniques, it is worthwhile to take a step back and ask:
\textit{1) Are the existing metrics truly capturing the hallucinations effectively? 2) Do these metrics generalize across different datasets, decoding techniques, model families, and model sizes?}
Confronting these questions is vital, as any attempt to alleviate hallucinations is futile unless we ensure its robust, reliable, and accurate measurement. 

The term \textit{`hallucination'} covers a spectrum of generation errors.
In this work, we focus on its two most common manifestations: poor \textit{faithfulness} and \textit{factuality}, particularly in knowledge-grounded dialog \cite{dziri-etal-2022-faithdial, dziri-etal-2022-evaluating} and question-answering \cite{lin-etal-2022-truthfulqa, li-etal-2023-halueval}. 
\textit{Faithfulness} measures the consistency and truthfulness with the provided knowledge source, while \textit{factuality} pertains to the accuracy with respect to 
real-world facts or widely accepted knowledge \cite{maynez-etal-2020-faithfulness}.

Measuring these constructs is no simple task.
In some cases, simple syntactic \cite{lin-2004-rouge} or sematic \cite{Zhang*2020BERTScore:} overlap with the input knowledge can provide an easy estimate of faithfulness.
Whereas other times, one has to resort to custom-trained models \cite{zhong-etal-2022-towards, dziri-etal-2022-faithdial}, multi-step question answering pipelines \cite{honovich-etal-2021-q2}, or LLM-based evaluation \cite{yan2024llm-evaluator, bavaresco2024llms}.
Interestingly, while recent surveys have extensively explored the causes and mitigation techniques for hallucinations in language models \cite{ji2023survey, zhang2023siren, chen2023hallucination, li2024dawn, huang2023survey}, none have directly called into question the generalization capabilities of existing metrics. 
Thus, in this work, we attempt to fill this gap, and conduct a rigorous empirical investigation of contemporary hallucination detection metrics.
Our study examines the above mentioned diverse sets of metrics from various perspectives -- consistency, alignment with human judgments, variation across decoding methods, impact of post-training, and the effect of parameter scaling. 

Our findings reveal that most metrics have limited inter-correlation and fail to consistently align with the human notion of hallucination. 
They seem to have a limited understanding of the problem, as they fail to generalize across datasets.
Anticlimactically, these metrics do not show a clear monotonic improvement with an increase in model size. 
On a positive note, we find that LLM-based evaluation, particularly with GPT-4, offers the most reliable detection across diverse tasks and datasets. 
Additionally, an ensemble of metrics also seems to be a good choice.
Instruction-tuning and mode-seeking decoding methods are also shown to reduce hallucinations.
We thus find that detecting hallucination does not have a \textit{one-size-fits-all} solution, as existing metrics fall short of capturing its full spectrum.
\section{Experimental Setup}
\label{sec:exp_setup}

\paragraph{Datasets.}
We focus on four datasets.
\underline{FaithDial} \cite{dziri-etal-2022-faithdial} and \underline{\textsc{Begin}} \cite{dziri-etal-2022-evaluating} are knowledge-grounded dialog datasets, where, given a conversation history \( H = (u_1, \dots, u_{n-1}) \) and knowledge source \(K_n\), the system generates a response \( \bar{u}_n \) that is coherent with \( H \) and supported by a non-empty subset \( M_n \subset K_n \) to be considered faithful.
\underline{TruthfulQA} \cite{lin-etal-2022-truthfulqa} is a factual question-answering dataset with multiple plausible answers. We measure factuality by comparing the generated answer's alignment with them.
Lastly, we analyze the knowledge-grounded QA and dialog subsets of the \underline{HaluEval} \cite{li-etal-2023-halueval} benchmark.
More details are provided in Appendix \S \ref{sec:datasets}.

\paragraph{Language Models.} 
Our study includes five LLM families: \underline{OPT} \cite{zhang2022opt}, \underline{Llama} \cite{touvron2023llama, dubey2024llama}, \underline{OLMo} \cite{groeneveld-etal-2024-olmo}, \underline{Phi} \cite{gunasekar2023textbooks, li2023textbooks, abdin2024phi}, and \underline{Gemma} \cite{team2024gemma}. 
We cover models ranging from 125M to 70B, including their instruction-tuned versions, totaling 37 models. 
Evaluation spans five decoding methods of greedy, beam search \cite{graves2012sequence}, ancestral, top-k \cite{fan-etal-2018-hierarchical}, and top-p sampling \cite{Holtzman2020The}.

\paragraph{Metrics.}
We evaluate hallucinations using the following six types of metrics.
1) \underline{Rouge-L} \cite{lin-2004-rouge}, \underline{Sacrebleu} \cite{post-2018-call}, and \underline{Knowledge-F1} measure the n-gram overlap between the generation, and reference text and source knowledge, respectively. 
2) Likewise, \underline{BertScore} \cite{Zhang*2020BERTScore:} and \underline{Knowledge-BertScore} \cite{dziri-etal-2022-faithdial, dziri-etal-2019-evaluating} assess their semantic similarity.
3) The pre-trained evaluator of consistency and groundedness from the \underline{UniEval suite} \cite{zhong-etal-2022-towards} help measure the factual alignment and input faithfulness, respectively.
4) \underline{$\mathcal{Q}^2$} \cite{honovich-etal-2021-q2} is a QA-based faithfulness metric that generates questions from the model output, identifies relevant spans in the knowledge source and ground truth response \cite{durmus-etal-2020-feqa, wang-etal-2020-asking}, and compares candidate answers to gold answers using either token-level or NLI-based F1.
5) \underline{Critic} \cite{dziri-etal-2022-faithdial} is an NLI-based classifier trained on dialog data, that identifies unfaithful responses.
\underline{GPT-4} \cite{achiam2023gpt} is used as an LLM-judge \cite{bavaresco2024llms}, that classifies hallucinated responses.
6) Finally, we combine consistency, K-BertScore, $\mathcal{Q}^2$ NLI, Critic, and GPT-4 scores using \textit{Factor Analysis of Mixed Data (FAMD)} \cite{pages2014multiple} to create an \underline{Ensemble} metric.
\section{Results and Discussion}
\label{sec:disc}

\begin{tcolorbox}[enhanced,attach boxed title to top center={yshift=-3mm,yshifttext=-1mm},
colback=red!20,colframe=black,colbacktitle=red!50,
title=Finding 1,boxrule=1pt,fonttitle=\bfseries\color{black},
boxed title style={size=small,colframe=black},left=1mm, right=1mm]
\centering
Except GPT-4, none of the metrics show consistent alignment with human judgment
\label{finding_1}
\end{tcolorbox}

\begin{table}[t]
\vspace{-3mm}
\begin{center}
\setlength{\tabcolsep}{3pt} 
\resizebox{\columnwidth}{!}{
\begin{tabular}{ccccccc}
\toprule
& \multicolumn{2}{c}{\textbf{Weighted-F1}} & \multicolumn{4}{c}{\textbf{PRAUC}}\\
\cmidrule(lr){2-3} 
\cmidrule(lr){4-7} 
\multirow{-2}{*}{\textbf{Dataset}} & \textbf{Critic} & \textbf{GPT-4} & \textbf{Consistency} & \textbf{K-BertScore} & \textbf{$\mathcal{Q}^2$ NLI} & \textbf{Ensemble}\\
\midrule
\textsc{Begin} CMU & $\textcolor{Tan}{\textbf{0.77}}$ & $\textcolor{ForestGreen}{\underline{\textbf{0.84}}}$ & $0.65$ & $0.70$ & $0.73$ & $0.65$\\
\textsc{Begin} TC & $\textcolor{Tan}{\textbf{0.74}}$ & $0.71$ & $0.65$ & $0.67$ & $\textcolor{ForestGreen}{\underline{\textbf{0.76}}}$ & $0.65$\\
\textsc{Begin} WoW & $\textcolor{Tan}{\textbf{0.83}}$ & $0.77$ & $0.43$ & $0.43$ & $0.56$ & $\textcolor{ForestGreen}{\underline{\textbf{0.96}}}$\\
\midrule
HaluEval Dial & $0.49$ & $\textcolor{ForestGreen}{\underline{\textbf{0.74}}}$ & $0.39$ & $\textcolor{Tan}{\textbf{0.61}}$ & $0.54$ & $ 0.42$\\
HaluEval QA & $0.53$ & $0.66$ & $0.36$ & $\textcolor{Tan}{\textbf{0.83}}$ & $0.82$ & $\textcolor{ForestGreen}{\underline{\textbf{0.93}}}$\\
\midrule
Average & $0.67$ & $\textcolor{ForestGreen}{\underline{\textbf{0.74}}}$ & $0.50$ & $0.65$ & $0.68$ & $\textcolor{Tan}{\textbf{0.72}}$\\
\bottomrule
\end{tabular}%
}
\end{center}
\vspace{-3mm}
\caption{Agreement between different metrics and human annotations. \textcolor{ForestGreen}{\underline{Green}} and \textcolor{Tan}{brown} denote the best and second-best metrics, respectively.
}
\vspace{-4mm}
\label{tab:all_dataset_auc}
\end{table}

Table \ref{tab:all_dataset_auc} displays the alignment scores of various metrics with human labels.
\footnote{Only \textsc{Begin} and HaluEval are included in this study, as they provide LLM-generated responses labeled for hallucinations.}.
Using PRAUC for continuous metrics and weighted-F1 for binary metrics (with random baseline scores of $0.50$ and $0.50-0.56$, respectively), we find mixed results across evaluation methods.
The UniEval suite's factual consistency evaluator performed just about at or below random chance across all the six data subsets.
K-BertScore and $\mathcal{Q}^2$ NLI both show strong performance on \textsc{Begin} CMU and HaluEval QA, with the latter also doing well on \textsc{Begin} TC. However, they both struggle  to replicate performance on \textsc{Begin} WoW and HaluEval Dial.
Critic, as expected, excels on the \textsc{Begin} benchmark, since it is trained on dialog datasets.
However, surprisingly, it drastically under performs on the HaluEval tasks, faring worse than even the random baseline.
The GPT-4 evaluator consistently shows agreeable alignment on average, acing on two datasets.
Our proposed ensemble metric is a close second, excelling particularly on the \textsc{Begin} WoW and HaluEval QA subsets.
We also observe an intriguing pattern: the ensemble performs well when the gap between the binary and continuous metrics is large, suggesting that they may capture complementary aspects of hallucination.
We discuss more metrics and related findings in Appendix \ref{sec:hyp_1}.
\begin{figure}[t]
    \centering
    \includegraphics[width=\columnwidth]{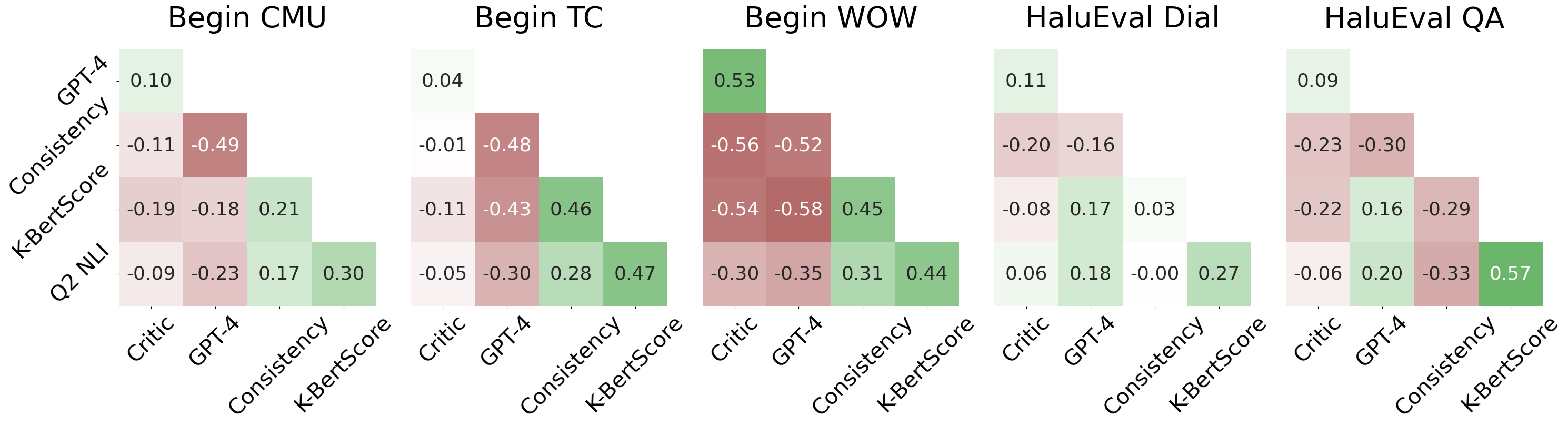}
    \vspace{-3mm}
    \caption{Spearman rank correlation between hallucination metrics reveals weak to no correlation for both \textsc{Begin} and HaluEval datasets.}
    \label{fig:begin_halueval_corr}
\end{figure}

\begin{figure}[t]
    \centering
    \includegraphics[width=\columnwidth]{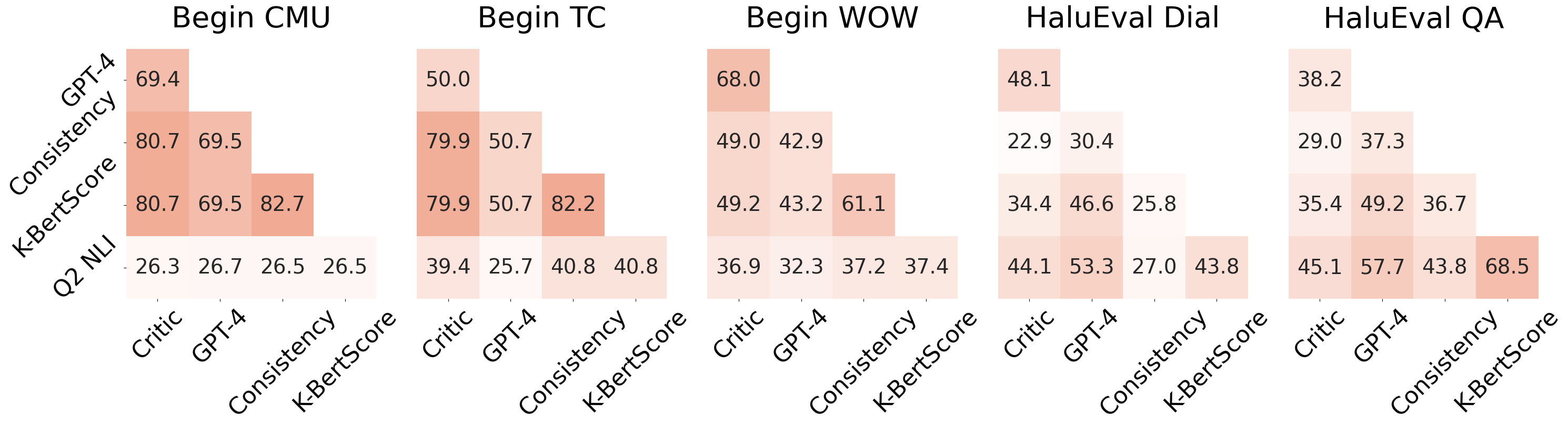}
    \vspace{-3mm}
    \caption{Percentage of correct matching labels shows minimal overlap between metrics' predictions.}
    \vspace{-5mm}
    \label{fig:begin_halueval_exact_match_gold}
\end{figure}

\vspace{-1mm}
\begin{tcolorbox}[enhanced,attach boxed title to top center={yshift=-3mm,yshifttext=-1mm},
colback=red!20,colframe=black,colbacktitle=red!50,
title=Finding 2,boxrule=1pt,fonttitle=\bfseries\color{black},
boxed title style={size=small,colframe=black},left=2mm, right=2mm]
\centering
Inter-metric correlation is weak
\label{finding_2}
\end{tcolorbox}

As shown in Figure \ref{fig:begin_halueval_corr}, the UniEval's consistency evaluator shows a moderate negative correlation with GPT-4 on the \textsc{Begin} datasets. In contrast, K-BertScore and $\mathcal{Q}^2$ NLI metrics exhibit a mild positive correlation across all datasets.
Interestingly, from Table \ref{tab:all_dataset_auc}, we see that Critic and GPT-4 produce similar results for \textsc{Begin} CMU and WoW, but their correlations differ significantly. 
These findings are consistent with TruthfulQA and FaithdDial, as shown in Figure \ref{fig:truthfulqa_faithdial_corr}.
Significant inter-metric agreement only appears in the \textsc{Begin} WoW corpus. 

To examine the differences in metric predictions, we plot the percentage overlap of their correct predictions in Figure \ref{fig:begin_halueval_exact_match_gold}.
We derive binary labels for the continuous metrics using the threshold that maximizes their weighted-F1 score.
The heatmap shows that the consistency evaluator and K-BertScore have over $80\%$ overlap for \textsc{Begin} CMU and TC. 
However, a closer look at the predicted label distribution (Table \ref{tab:begin_halueval_metric_label_dist} in \S \ref{sec:hyp_1}) reveals that they always classify generations as hallucinations, indicating their limited understanding of the construct.
Moreover, because of the skewed label distribution of \textsc{Begin} CMU and TC, these metrics' predictions largely overlap with those of more accurate metrics like Critic and GPT-4, creating a false \textit{mirage} of their success.
The latter also demonstrate high overlap with each other on the \textsc{Begin} datasets. 
$\mathcal{Q}^2$ NLI shows minimal overlap with other metrics, except for K-BertScore in HaluEval QA -- the only instance where both perform well. 
Otherwise, all other metrics show little overlap.


\begin{tcolorbox}[enhanced,attach boxed title to top center={yshift=-3mm,yshifttext=-1mm},
colback=ForestGreen!20,colframe=black,colbacktitle=ForestGreen!50,
title=Finding 3,boxrule=1pt,fonttitle=\bfseries\color{black},
boxed title style={size=small,colframe=black},left=1mm, right=1mm]
\centering
Instruction-tuning and mode-seeking decoding methods reduce hallucinations
\label{finding_3}
\end{tcolorbox}

\begin{table}[t]
\vspace{-3mm}
\centering
\setlength{\tabcolsep}{3pt} 
\resizebox{0.8\columnwidth}{!}{%
\begin{tabular}{ccccccc}
\toprule
\multirow{3}{*}{\textbf{Metric}} & \multicolumn{3}{c}{\textbf{TruthfulQA}} & \multicolumn{3}{c}{\textbf{FaithDial}} \\
\cmidrule(lr){2-4} 
\cmidrule(lr){5-7} 
& \textbf{Training} & \textbf{Model} & \textbf{Decoding} & \textbf{Training} & \textbf{Model} & \textbf{Decoding} \\
& \textbf{Type} & \textbf{Size} & \textbf{Method} & \textbf{Type} & \textbf{Size} & \textbf{Method} \\

\midrule
Rouge-L & 0.0 & 0.028 & 0.0 & 0.013 & 0.0 & 0.0 \\
Sacrebleu & 0.0 & 0.0 & 0.0 & \cellcolor{red!20}0.246 & 0.0 & 0.0 \\
BertScore & 0.0 & \cellcolor{red!20}0.218 & 0.0 & 0.0 & 0.001 & 0.0 \\
Groundedness & 0.0 & 0.01 & 0.0 & 0.0 & 0.0 & 0.0 \\
\midrule
Consistency & 0.01 & 0.0 & \cellcolor{red!20}0.207 & 0.03 & \cellcolor{red!20}0.489 & 0.0 \\
K-BertScore & 0.0 & \cellcolor{red!20}0.120 & 0.0 & \cellcolor{red!20}0.116 & 0.0 & 0.0 \\
$\mathcal{Q}^2$ NLI & 0.0 & 0.0 & 0.0 & 0.005 & 0.012 & 0.0 \\
Critic & 0.0 & 0.0 & 0.0 & 0.013 & \cellcolor{red!20}0.289 & 0.0 \\
GPT-4 & 0.0 & 0.0 & 0.0 & 0.0 & 0.0 & 0.0 \\
\bottomrule
\end{tabular}%
}
\caption{Significance test results for the impact of training type, model size, and decoding methods on hallucination metrics. \colorbox{red!20}{Red cells} ($p > 0.05$) indicate failure to reject the null hypothesis.}
\label{tab:p_val}
\vspace{-5mm}
\end{table}

\begin{figure*}[!ht]
    \centering
    \includegraphics[width=0.9\textwidth, height=4cm]{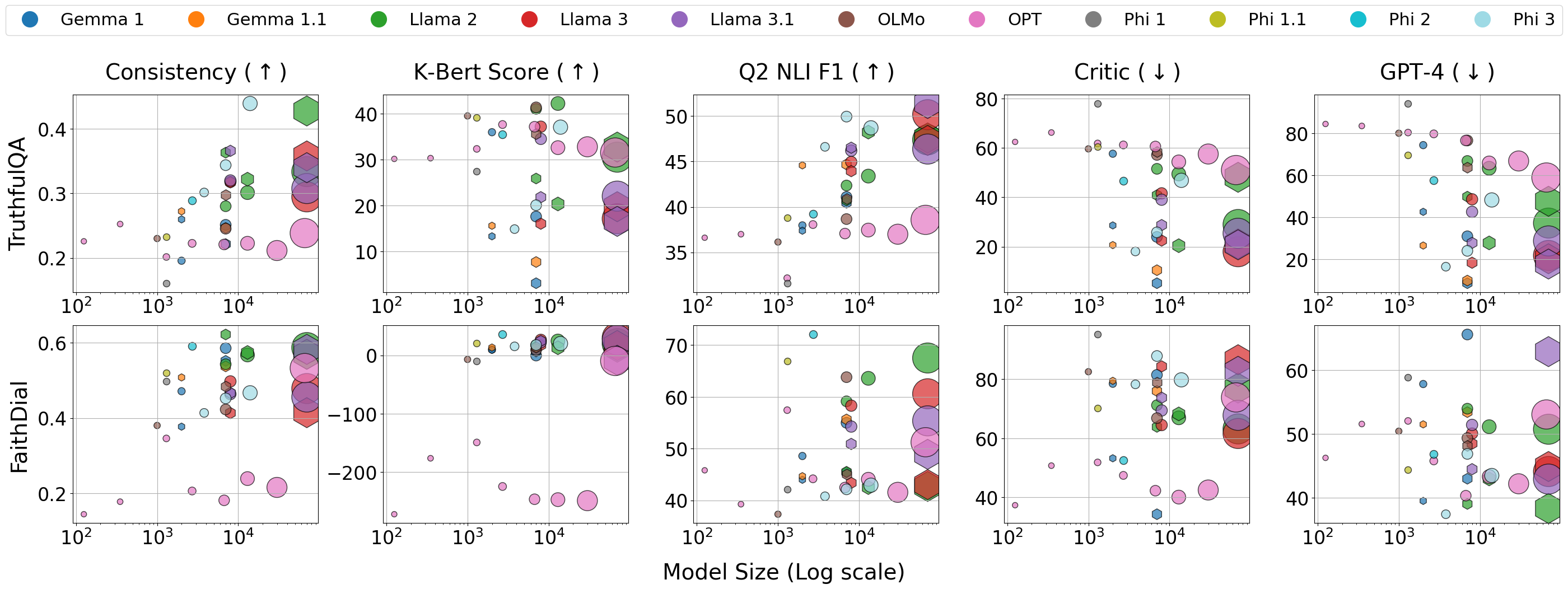}
    \caption{Hallucination detection metric scores for greedy decoding on various model sizes. Circles and hexagons represent pretrained and instruction-tuned models, respectively. 
    }
    \label{fig:scale_fact}
\end{figure*}

\begin{figure}[t]
    \centering
    \includegraphics[width=\columnwidth,height=4.5cm]{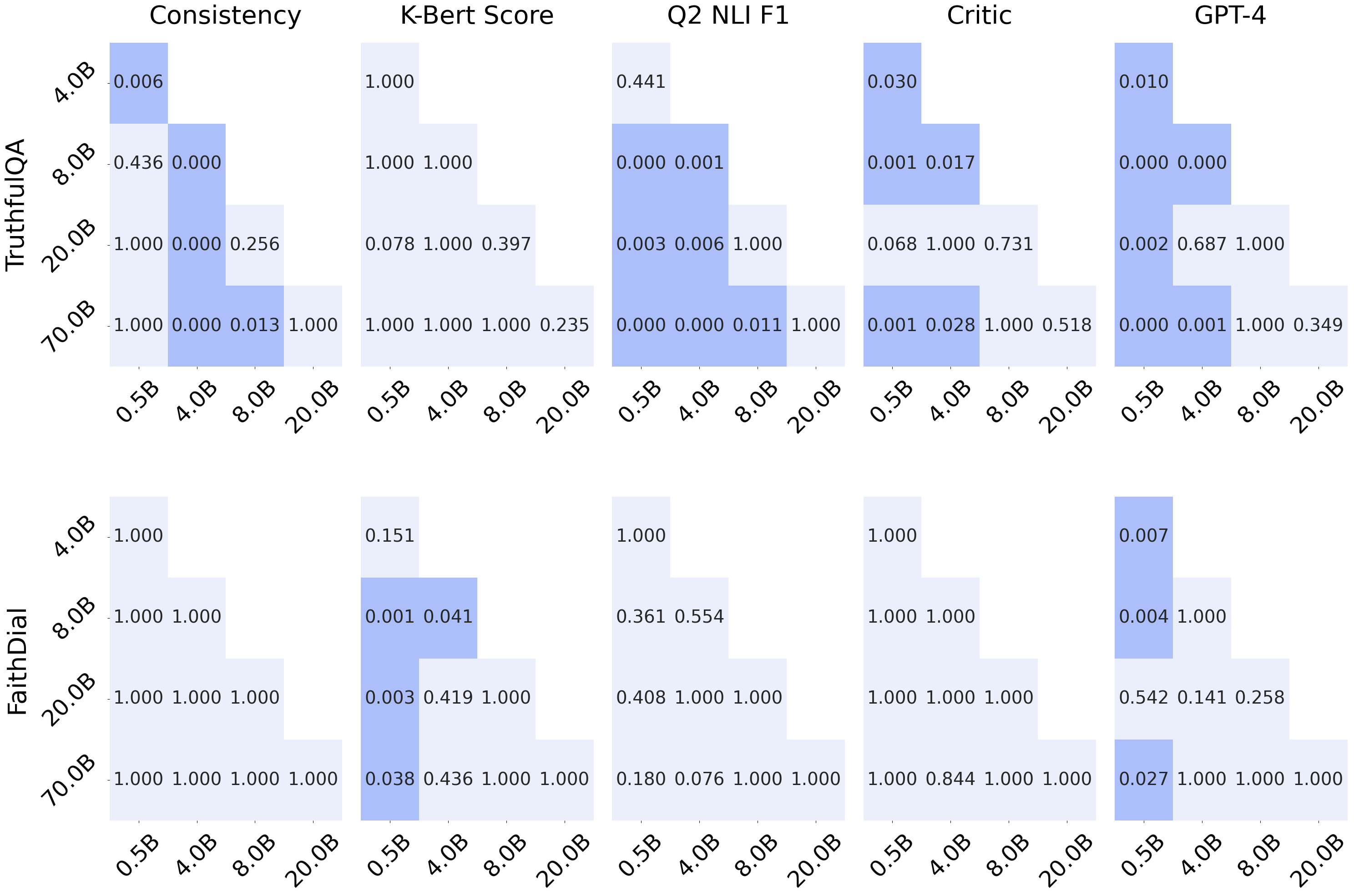}
    \caption{P-values for different model size bins from the pairwise Mann-Whitney rank test.}
    \label{fig:size_hyp_testing_fact}
\end{figure}

Instruction-tuning is known to perform well on grounded generation tasks, and to reduce hallucinations \cite{ouyang2022training, dhuliawala2023chain, kalai2024calibrated}.
To revalidate these findings, we analyze TruthfulQA and FaithDial, conducting paired significance tests (detailed in Appendix \S \ref{sec:hyp_testing}) on various hallucination metrics between pre-trained models and their instruction-tuned versions from \S \ref{sec:exp_setup}.
The null hypothesis posits that `\textit{Instruction-tuning has no effect on hallucination detection metrics}'.
The results in Table \ref{tab:p_val} help us refute this claim, albeit with some exceptions -- SacreBleu and K-BertScore show no significant gains with instruction-tuning on FaithDial.
Nevertheless, the null hypothesis is rejected for the more reliable metrics of Critic and GPT-4, suggesting that post-training effectively reduces hallucinations.
Shifting focus to decoding techniques, it is well established that mode-seeking decoding methods such and greedy and beam search tend to hallucinate less than sampling methods (ancestral, top-p, and top-k) \cite{dziri-etal-2021-neural, li2024dawn}. 
Our paired significance test results in Table \ref{tab:p_val} confirm these findings. Additionally, the posthoc pairwise significance testing results in Figures \ref{fig:decoding_hyp_testing_fact} and \ref{fig:decoding_hyp_testing_nlg} (Appendix \S \ref{sec:extended_hyp_3}) strengthen our argument.


\begin{tcolorbox}[enhanced,attach boxed title to top center={yshift=-3mm,yshifttext=-1mm},
colback=red!20,colframe=black,colbacktitle=red!50,
title=Finding 4,boxrule=1pt,fonttitle=\bfseries\color{black},
boxed title style={size=small,colframe=black},left=1mm, right=1mm]
\centering
Metrics do not show commensurate gains with parameter scaling
\label{finding_4}
\end{tcolorbox}

Scaling language model parameters typically leads to a monotonic increase in both pretraining \cite{kaplan2020scaling, hoffmann2022training} and downstream metrics \cite{caballero2023broken}, often following a power law. 
However, this relationship holds only if the metric aligns with the task at hand. 
Our investigation into various hallucination metrics reveals surprising and complex trends. 
As seen in Figure \ref{fig:scale_fact}, no clear linear or monotonic patterns emerge across the metrics for both datasets. 
Critic also shows contradictory trends in TruthfulQA and Faithdial.
Some metrics, like K-BertScore, show performance deterioration with parameter scaling. 
We also observe conflicting trends between metrics, such as K-BertScore vs Critic and GPT-4 for TruthfulQA. 
The results for Gemma 1 and 1.1 often suggest opposite conclusions regarding hallucinations.
Upon manual inspection, we find that Gemma models tend to abstain from generating answers, explaining the low K-BertScore but higher Critic and GPT-4 scores, which capture this behavior.
Similar underperformance trends are evident across other metric types, as shown in Figures \ref{fig:scale_nlg} and \ref{fig:scale_other}. 
More findings are presented in Appendix \S \ref{sec:extended_hyp_4}.

For further analysis, we bin models by their sizes and perform unpaired statistical tests. 
The null hypothesis here is that `\textit{Parameter scaling has no effect on metric performance}'.
As shown in Table \ref{tab:p_val}, only GPT-4 consistently rejects the null hypothesis, indicating that it is the only metric whose performance improves with an increase in model size. 
Figure \ref{fig:size_hyp_testing_fact} shows posthoc pairwise p-values. $\mathcal{Q}^2$ NLI and Critic for FaithDial, and K-BertScore for TruthfulQA, show little improvement with parameter scaling.
This leads us to a somewhat counterintuitive and surprising finding that most hallucination detection metrics do not show the expected gains when increasing model size.
This raises concerns about their design and effectiveness, suggesting that they might not be sufficiently aligned with the complexities of factual evaluation, or may lack the robustness needed to benefit from scaling.

\section{Failure Modes of the Metrics}

\begin{figure*}[!ht]
    \centering
    \includegraphics[width=\textwidth, height=4cm]{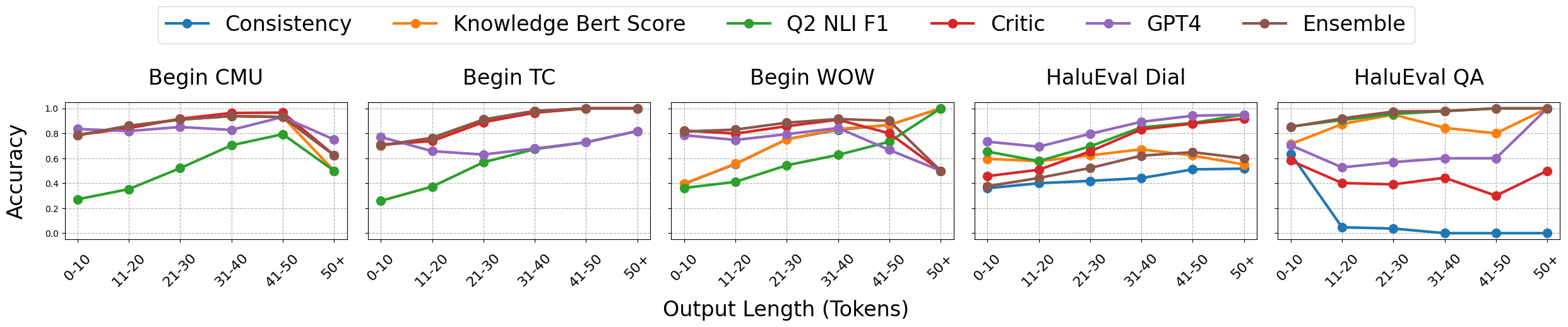}
    \caption{Metric accuracy across varying response lengths on the \textsc{Begin} and HaluEval. 
    }
    \label{fig:error_analysis}
    \vspace{-5mm}
\end{figure*}

To better understand the limitations and bottlenecks of evaluation metrics, we analyze how their performance varies with respect to instance difficulty.
While `\textit{difficulty}' can be quantified in various ways, it is well-established that LLMs tend to struggle and exhibit increased hallucination rates, with longer responses \cite{yang2025hallucinate, lauscher2025much}. 
Thus, we investigate the potency of different metrics across varying response lengths. 
Specifically, we calculate the accuracy of each metric for different response lengths, as shown in Figure \ref{fig:error_analysis}. 
Appendix \ref{sec:extended_discussion} contains additional error analysis and discussion.

\noindent\textbf{UniEval suite's Consistency} performs reasonably well on longer responses in the \textsc{Begin} CMU and \textsc{Begin} TC datasets, where accuracy rises steadily with response length -- reaching values above $0.9$.
However, its performance on \textsc{Begin} WOW starts quite low ($0.4$ in short completions), and only becomes competitive with length. 
Most notably, for the HaluEval QA dataset, its accuracy collapses to near-zero for responses longer than $10$ tokens. 
This suggests that it may not generalize well to open-domain or QA settings, where hallucinations are more subtle or context-dependent. 

\noindent\textbf{K-BERTScore} exhibits similar trends -- it improves with response length and performs best on datasets like \textsc{Begin} CMU, \textsc{Begin} TC, and HaluEval Dial, even reaching perfect accuracy in some of the longer bins. 
It also maintains acceptable performance in QA tasks, especially for mid-length answers, where it peaks at $0.95$. 
However, its reliability drops for very short outputs. 
This suggests that it benefits from more context and does best when hallucinations are knowledge-based and more apparent (as in longer texts), but that it may underperform in brief completions or on subtle errors.  

\noindent\textbf{$\mathcal{Q}^2$ NLI} stands out in the QA task. 
It consistently achieves high accuracy ($\geq 0.85$), going up to $1.0$ in longer bins. 
Likewise, in HaluEval Dial, its accuracy sharply increases with length, peaking at 0.95 in the longest bin. 
However, it falters on the \textsc{Begin} datasets, especially for shorter completions, where it struggles to surpass $0.4$. 
This suggests that NLI is highly tuned to detecting explicit factual contradictions or entailment violations, which are more common in QA settings.

\noindent\textbf{Critic} displays one of the most balanced performance profiles across the dialog datasets. 
In both \textsc{Begin} CMU and \textsc{Begin} TC, its accuracy consistently improves with output length, reaaching near-perfect values in the longer bins (e.g., $0.97$). 
It is also strong in WOW and HaluEval Dial, maintaining $> 0.8$ accuracy for nearly all lengths. 
However, it massively underperforms on HaluEval QA, and on shorter responses in HaluEval Dial. 
This shows that Critic is well-suited for dialogues and is relatively robust to variation in output length, but might not be the best choice for QA. 

\noindent\textbf{GPT-4} performs strongly overall, especially on longer outputs. 
For most datasets, it achieves accuracy $ > 0.75$ across different response lengths, often peaking above $0.9$. 
However, for \textsc{Begin} TC, its accuracy is unexpectedly lower ($\sim0.65$) for mid-length completions. 
Likewise, for \textsc{Begin} WOW or HaluEval QA, short responses sometimes yield surprisingly high or low scores. 
Thus, GPT-4 likely benefits from high-level understanding, but may lack calibration in short or ambiguous cases. 

\noindent\textbf{Ensemble} yields strong performance across all datasets and lengths, and almost always performs on par with or better than the best individual metric.
It reaches near-perfect accuracy for HaluEval QA. 
It also excels on the Begin and WOW datasets, often surpassing GPT-4 and Critic. 
Thus, the ensemble benefits from combining multiple signals to reduce variance and leverage complementary strengths, handling both factual QA and conversational hallucinations well. 
\section{Conclusion}

Hallucination detection in LLM-generated text is a tricky task. 
Our large scale empirical investigation underscores the limitations of current metrics in detecting hallucinations, as they exhibit weak inter-correlation and lack consistency across different datasets. 
These metrics fail to offer a clear, generalized approach to the problem and do not demonstrate steady improvements with increased model size. 
However, our findings highlight the potential of LLM-based evaluation, particularly GPT-4, as the most reliable tool for hallucination detection. Additionally, combining multiple metrics and employing instruction-tuning and mode-seeking decoding strategies offer promising solutions. 
Ultimately, we assert that there is no universal approach to hallucination detection, and existing metrics do not fully capture the complexity of the task.\looseness=-1

\section{Limitations}
While our large-scale empirical investigation provides a thorough analysis of the current hallucination metrics, it does have certain limitations.

\paragraph{Evaluation Setup}
As noted in \S\ref{sec:exp_setup}, the \textsc{begin} and HaluEval datasets include model-generated responses with human annotations for hallucinations, whereas TruthfulQA and FaithDial do not.
We therefore generate responses for the latter using the models described in \S\ref{sec:exp_setup}. 
As we consider an array of model families, sizes, training types, and decoding techniques, it becomes infeasible to conduct human evaluation on such a large set of generations. 
Consequently, for Finding 1, we focus solely on the \textsc{begin} and HaluEval. 
Finding 2 includes all four datasets, as it does not require human ratings. 
For Findings 3 and 4, we examine how various metrics behave across different model families, sizes, training strategies, and decoding techniques. 
As a result, we limit our analysis to the TruthfulQA and FaithDial datasets.
Additionally, since FaithDial is a modified version of the WoW dataset \cite{dinan2018wizard}, which is already included in \textsc{begin}, we can reasonably assume that the Findings 3 and 4 results for \textsc{begin} will follow similar trends to those observed for FaithDial.

\paragraph{Tasks and Datasets.}
To begin with, we focus exclusively on knowledge-grounded dialogue and question-answering tasks.
However, hallucination is a prevalent issue across various other NLP tasks, such as machine translation \cite{raunak-etal-2021-curious}, summarization \cite{cao-etal-2022-hallucinated}, code-generation \cite{liu2024exploring}, and linguistic applications \cite{weissweiler-etal-2023-counting}. 
Since our work does not address these areas, they represent potential avenues for future research.

\paragraph{Evaluation Metrics.}
Existing research on detecting hallucinations in LLM-generated text typically includes statistical, model-based, and human-based evaluations \cite{ji2023survey, chen2023hallucination}. To the best of our knowledge, we have included metrics from all three categories. ROUGE and BERTScore capture statistical overlap, while the UniEval suite, $\mathcal{Q}^2$, Critic, and GPT-4-as-judge are model-based metrics. The only other remaining category consists of uncertainty-based metrics, such as Semantic Entropy \cite{kuhn2023semantic} and SAR \cite{duan-etal-2024-shifting} among others. However, these methods require access to token log probabilities, which are not available in the \textsc{BEGIN} and HaluEval datasets, preventing us from evaluating their alignment with human judgments. Furthermore, while we identify LLM-as-judge as the most reliable metric for hallucination detection, we do not evaluate its variants -- such as chain-of-thought prompting \cite{akbar-etal-2024-hallumeasure}, \textsc{G-eval} \cite{liu-etal-2023-g}, or smaller / different architecture LLMs \cite{thakur2024judging} -- due to the scope of our study. 
Lastly, while fine-tuning has been shown to mitigate model hallucinations \cite{ghosal2024understanding}, we have not explored these experiments in our study, leaving them for future investigation.

\bibliography{custom}

\clearpage
\newpage
\appendix
\section{Appendix}
\label{sec:appendix}
\label{sec:halu_def}

\subsection{Datasets}
\label{sec:datasets}

\begin{itemize}[leftmargin=*]
    \item \textbf{\textsc{Begin} Benchmark \cite{dziri-etal-2022-evaluating}:} It is a collection of 3 knowledge-grounded dialog datasets : CMU-Dog \cite{zhou-etal-2018-dataset}, Wizard of Wikipedia (WoW) \cite{dinan2018wizard}, and TopicalChat \cite{gopalakrishnan19_interspeech}. 
    It contains responses generated by 4 models : GPT2 \cite{Radford2019LanguageMA}, T5 \cite{raffel2020exploring}, DoHA (BART with dual attention) \cite{prabhumoye-etal-2021-focused}, and CTRL-T5 (control tokens augmented T5) \cite{rashkin-etal-2021-increasing}. 
    Each response is also annotated as one among \textit{faithful}, \textit{unfaithful}, or \textit{generic} by human annotators.
    For all our experiments, we ignore the instances that were labeled \textit{generic}. 
    We analyze the metrics listed in Section \S \ref{sec:exp_setup} using the responses provided and annotated in the dataset, rather than by generating new responses.
    
    \item \textbf{HaluEval \cite{li-etal-2023-halueval}:} It is a conglomerate of $5,000$ general-purpose and $30,000$ task-specific examples designed for hallucination evaluation, spanning question answering, knowledge-grounded dialog, and text summarization. 
    We focus on the task-specific subset, which includes $10,000$ examples randomly sampled from the training sets of HotpotQA \cite{yang-etal-2018-hotpotqa}, and OpenDialKG \cite{moon-etal-2019-opendialkg}.
    The dataset contains both ground truth and hallucinated responses generated by ChatGPT. 
    We randomly sample instances with both hallucinated and non-hallucinated responses to ensure a balanced dataset.
    As with the \textsc{Begin} benchmark setup, we analyze the responses released by the HaluEval authors, rather than generating new ones.

    \item \textbf{TruthfulQA \cite{lin-etal-2022-truthfulqa}:} It assess how accurately a language model generates truthful answers to factual questions. 
    It features $817$ questions across $38$ diverse categories. 
    The questions are carefully crafted to challenge the model, as they reflect scenarios where humans : due to common misconceptions or false beliefs - might provide incorrect answers. 
    It contains two setups : text generation and multiple choice questions. 
    We consider the former in our experiments. 
    We generate the answers using the models mentioned in \S \ref{sec:exp_setup}.

    \item \textbf{FaithDial \cite{dziri-etal-2022-faithdial}:} It is an information-seeking, knowledge-grounded dialog response generation dataset, that contains $50,000$ dialog turns spread across $5,500$ conversations. 
    It is curated by refining and correcting the hallucinated responses found in the \textit{Wizard-of-Wikipedia (WoW)} corpus \cite{dinan2018wizard}. 
    We generate the responses using the models mentioned in \S \ref{sec:exp_setup}. 
\end{itemize}

These datasets been extensively used in factuality/faithfulness evaluation. 
For instance, \textsc{Begin} and FaithDial have been studied in various works such as \citet{das-etal-2022-diving}, \citet{zha-etal-2023-alignscore}, and \citet{adlakha-etal-2024-evaluating} to name a few. 
Similarly, HaluEval has been used for factuality evaluation in the works of \citet{zhang2023alleviating} and \citet{xu2024sayself}).

Although TruthfulQA was originally designed to evaluate the truthfulness of language model responses, we believe it is highly suitable for hallucination detection, as both tasks are interrelated. 
The dataset includes questions like ``\textit{What happens if you eat watermelon seeds?}'' and ``\textit{Can coughing effectively stop a heart attack?}'' -- questions where incorrect answers from a language model can clearly be classified as `\textit{factual}' hallucinations. 
Additionally, other studies (e.g., \citet{zhang2024truthx} and \citet{du2024haloscope}) have successfully used TruthfulQA for hallucination detection, further validating its effectiveness for this purpose.

\subsection{Models}
\label{sec:models}
We use the following models in our experiments:
\begin{itemize}[leftmargin=*]
  \setlength\itemsep{1mm}
    \item \textbf{OPT \cite{zhang2022opt}:} 125M, 350M, 1.3B, 2.7B, 6.7B, 13B, 30B, 66B
    \item \textbf{Llama 2 \cite{touvron2023llama}:} 7B, 13B, 70B, 7B-chat, 13B-chat, and 70b-chat
    \item \textbf{Llama 3 and 3.1 \cite{dubey2024llama}:} 3B, 70B, 3B-Instruct, 70B-Instruct
    \item \textbf{Phi \cite{gunasekar2023textbooks, li2023textbooks, abdin2024phi}:} Phi-3-small-8k-instruct, Phi-3-mini-4k-instruct, Phi-3-medium-4k-instruct
    \item \textbf{Gemma \cite{team2024gemma}:} gemma-2b, gemma-7b, gemma-2b-it, gemma-7b-it, gemma-1.1-2b-it, gemma-1.1-7b-it
    \item \textbf{OLMo \cite{groeneveld-etal-2024-olmo}:} 1B, 7B, 1B-Instruct, and 7B-Instruct
\end{itemize}

\begin{table*}[!ht]
    \centering
    \renewcommand{\arraystretch}{1.4}
    \small
    \begin{tabularx}{\textwidth}{X}
        \toprule
        \rowcolor{blue!10} You are comparing whether the submitted response is conditioned on the dialogue history and knowledge snippet. Here is the data: \\ 
        \rowcolor{blue!10} \texttt{[\textsc{Begin} DATA]} \\
        \rowcolor{blue!10} ************ \\
        \rowcolor{blue!10} \texttt{[Knowledge]}: \{knowledge\} \\
        \rowcolor{blue!10} ************ \\
        \rowcolor{blue!10} \texttt{[Dialog History]}: \{history\} \\
        \rowcolor{blue!10} ************ \\
        \rowcolor{blue!10} \texttt{[Submission]}: \{generated response\} \\
        \rowcolor{blue!10} ************ \\
        \rowcolor{blue!10} \texttt{[END DATA]} \\
        \rowcolor{blue!10} Compare the factual content of the submitted response with the knowledge snippet and consistency with the dialog history. Ignore any differences in style, grammar, or punctuation. \\
        \rowcolor{blue!10} The submitted response may either be a subset or superset of the knowledge snippet, or it may conflict with it. Determine which case applies. Answer the question by selecting one of the following options: \\
        \rowcolor{green!10} (A) The submitted response is a subset of the knowledge snippet and is fully consistent with the dialog history. \\
        \rowcolor{red!10} (B) The submitted response is a superset of the knowledge snippet and is fully consistent with the dialog history. \\
        \rowcolor{green!20} (C) The submitted response contains all the same details as the knowledge snippet and is fully consistent with the dialog history. \\
        \rowcolor{red!20} (D) There is a disagreement between the submitted response and the knowledge snippet or it is inconsistent with the the dialog history. \\
        \rowcolor{green!20} (E) The answers differ, but these differences don't matter from the perspective of factuality. \\
        \rowcolor{blue!10} Answer the question by calling \texttt{`select\_choice'} with your reasoning in a step-by-step matter to be sure that your conclusion is correct. Avoid simply stating the correct answer at the outset. \\
        \rowcolor{blue!10} Select a single choice by setting the \texttt{`choice'} parameter to a single choice from A, B, C, D, or E. \\
    \bottomrule
    \end{tabularx}
    \caption{GPT-4 Evaluator prompt for \textsc{Begin} benchmark.}
    \label{tab:begin_prompt}
\end{table*}

\begin{table*}[!ht]
    \centering
    \renewcommand{\arraystretch}{1.4}
    \small
    \begin{tabularx}{\textwidth}{X}
        \toprule
        \rowcolor{blue!10} You are comparing a submitted answer to a set of possible correct answers for a given question. Here is the data: \\ 
        \rowcolor{blue!10} \texttt{[\textsc{Begin} DATA]} \\
        \rowcolor{blue!10} ************ \\
        \rowcolor{blue!10} \texttt{[Question]}: \{question\} \\
        \rowcolor{blue!10} ************ \\
        \rowcolor{blue!10} \texttt{[Possible Correct Answers]}: \{correct answers\} \\
        \rowcolor{blue!10} ************ \\
        \rowcolor{blue!10} \texttt{[Submission]}: \{generated answer\} \\
        \rowcolor{blue!10} ************ \\
        \rowcolor{blue!10} \texttt{[END DATA]} \\
        \rowcolor{blue!10} Compare the factual content of the submitted answer with the set of possible correct answers. Ignore any differences in style, grammar, or punctuation. \\
        \rowcolor{blue!10} The submitted answer may either be a subset or superset of possible correct answers, or it may conflict with it. Determine which case applies. Answer the question by selecting one of the following options: \\
        \rowcolor{green!10} (A) The submitted answer is a subset of the possible correct answers and is fully consistent with it.\\
        \rowcolor{red!10} (B) The submitted answer is a superset of the possible correct answers and is fully consistent with it. \\
        \rowcolor{green!20} (C) The submitted answer contains all the same details as one of the possible correct answer. \\
        \rowcolor{red!20} (D) There is a disagreement between the submitted answer and all of the possible correct answers. \\
        \rowcolor{green!20} (E) The answers differ, but these differences don't matter from the perspective of factuality. \\
        \rowcolor{blue!10} Answer the question by calling \texttt{`select\_choice'} with your reasoning in a step-by-step matter to be sure that your conclusion is correct. Avoid simply stating the correct answer at the outset. \\
        \rowcolor{blue!10} Select a single choice by setting the \texttt{`choice'} parameter to a single choice from A, B, C, D, or E. \\
    \bottomrule
    \end{tabularx}
    \caption{GPT-4 Evaluator prompt for TruthfulQA benchmark.}
    \label{tab:truthfulqa_prompt}
\end{table*}

\begin{table*}[!ht]
    \centering
    \renewcommand{\arraystretch}{1.4}
    \small
    \begin{tabularx}{\textwidth}{X}
        \toprule
        \rowcolor{blue!10} You are comparing a submitted response to an expert response conditioned on a dialogue history and knowledge snippet. Here is the data: \\ 
        \rowcolor{blue!10} \texttt{[\textsc{Begin} DATA]} \\
        \rowcolor{blue!10} ************ \\
        \rowcolor{blue!10} \texttt{[Knowledge]}: \{Knowledge\} \\
        \rowcolor{blue!10} ************ \\
        \rowcolor{blue!10} \texttt{[Dialog History]}: \{history\} \\
        \rowcolor{blue!10} ************ \\
        \rowcolor{blue!10} \texttt{[Expert]}: \{gold response\} \\
        \rowcolor{blue!10} ************ \\
        \rowcolor{blue!10} \texttt{[Submission]}: \{generated response\} \\
        \rowcolor{blue!10} ************ \\
        \rowcolor{blue!10} \texttt{[END DATA]} \\
        \rowcolor{blue!10} Compare the factual content of the submitted response with the expert response and knowledge snippet. Ignore any differences in style, grammar, or punctuation. \\
        \rowcolor{blue!10} The submitted answer may either be a subset or superset of the expert response, or it may conflict with it. Determine which case applies. Answer the question by selecting one of the following options: \\
        \rowcolor{green!10} (A) The submitted response is a subset of the expert response and is fully consistent with it. \\
        \rowcolor{red!10} (B) The submitted response is a superset of the expert response and is fully consistent with it. \\
        \rowcolor{green!20} (C) The submitted response contains all the same details as the expert response. \\
        \rowcolor{red!20} (D) There is a disagreement between the submitted response and the expert response. \\
        \rowcolor{green!20} (E) The response differ, but these differences don't matter from the perspective of factuality. \\
        \rowcolor{blue!10} Answer the question by calling \texttt{`select\_choice'} with your reasoning in a step-by-step matter to be sure that your conclusion is correct. Avoid simply stating the correct answer at the outset. \\
        \rowcolor{blue!10} Select a single choice by setting the \texttt{`choice'} parameter to a single choice from A, B, C, D, or E. \\
    \bottomrule
    \end{tabularx}
    \caption{GPT-4 Evaluator prompt for FaithDial benchmark.}
    \label{tab:faithdial_prompt}
    \vspace{-3mm}
\end{table*}

\subsection{Prompts}
\label{sec:prompts}

The GPT-4 evaluator prompts for Begin, TruthfulQA, and FaithDial are outlined in Tables \ref{tab:begin_prompt}, \ref{tab:truthfulqa_prompt}, and \ref{tab:faithdial_prompt}, respectively. We use the OpenAI's \texttt{gpt-4o-mini} model.
The evaluator selects from the options $\{\texttt{A, B, C, D, E}\}$, with options $\texttt{B}$ and $\texttt{D}$ identified as hallucinated responses, while the others are deemed benign. 
These prompts are based on the factuality template from the Autoevals library\footnote{\url{https://github.com/braintrustdata/autoevals/}}.
For HaluEval, we leverage the pre-existing templates provided by \citet{li-etal-2023-halueval}.

\subsection{Hypothesis Testing}
\label{sec:hyp_testing}
We conduct various types of significance tests to support our findings from \S \ref{sec:disc}. 
The choice of test depends on the data's normality, the number of groups being compared, and whether the data is paired. Tables \ref{tab:size_hyp_table}, \ref{tab:training_type_hyp_table}, and \ref{tab:decoding_hyp_table} detail the different tests used for our experiments.

\begin{table}[!ht]
\begin{center}
\small
\setlength{\tabcolsep}{3pt} 
\resizebox{\columnwidth}{!}{
\begin{tabular}{@{\hskip 0pt} >{\centering\arraybackslash}p{0.2\columnwidth} >{\centering\arraybackslash}p{0.4\columnwidth} >{\centering\arraybackslash}p{0.4\columnwidth} @{\hskip 0pt}} 
\toprule
\textbf{Test} & \textbf{TruthfulQA} & \textbf{Faithdial} \\
\midrule
Dependent T-Test & K-BertScore, $\mathcal{Q}^2 $ NLI  & RougeL, Sacrebleu \\
\midrule
Wilcoxon Signed-Rank Test & RougeL, Sacrebleu, BertScore, Groundedness, Consistency, Critic, GPT-4 & BertScore, Groundedness, Consistency, K-BertScore, $\mathcal{Q}^2 $ NLI, Critic, GPT-4\\
\bottomrule
\end{tabular}
}
\end{center}
\caption{Hypothesis tests comparing instruction tuning vs pretraining: Dependent T-Test for normal data, Wilcoxon Signed-Rank Test otherwise.}
\label{tab:training_type_hyp_table}
\end{table}

\begin{table}[!ht]
\begin{center}
\small
\setlength{\tabcolsep}{3pt} 
\resizebox{\columnwidth}{!}{
\begin{tabular}{@{\hskip 0pt} >{\centering\arraybackslash}p{0.2\columnwidth} >{\centering\arraybackslash}p{0.35\columnwidth} >{\centering\arraybackslash}p{0.45\columnwidth} @{\hskip 0pt}} 
\toprule
\textbf{Test} & \textbf{TruthfulQA} & \textbf{Faithdial} \\
\midrule
Repeated Anova Test & RougeL, K-BertScore, $\mathcal{Q}^2 $ NLI &  GPT-4 \\
\midrule
Friedman Test & Sacrebleu, BertScore, Groundedness, Consistency, Critic, GPT-4 & RougeL, Sacrebleu, BertScore, Groundedness, Consistency, K-BertScore, $\mathcal{Q}^2 $ NLI, Critic \\
\bottomrule
\end{tabular}
}
\end{center}
\caption{Hypothesis tests comparing decoding methods: Repeated Anova for normal data, Friedman Test otherwise, with Pairwise T-Tests (Bonferroni) for the former and Nemenyi test for the latter in posthoc analysis.}
\label{tab:decoding_hyp_table}
\end{table}

\begin{table}[!ht]
\begin{center}
\small
\setlength{\tabcolsep}{3pt} 
\resizebox{\columnwidth}{!}{
\begin{tabular}{@{\hskip 0pt} >{\centering\arraybackslash}p{0.2\columnwidth} >{\centering\arraybackslash}p{0.4\columnwidth} >{\centering\arraybackslash}p{0.4\columnwidth} @{\hskip 0pt}} 
\toprule
\textbf{Test} & \textbf{TruthfulQA} & \textbf{Faithdial} \\
\midrule
One-Way ANOVA Test & - &  GPT-4 \\
\midrule
Kruskal-Wallis & RougeL, Sacrebleu, BertScore, Groundedness, Consistency, K-BertScore, $\mathcal{Q}^2 $ NLI, Critic, GPT-4 & RougeL, Sacrebleu, BertScore, Groundedness, Consistency, K-BertScore, $\mathcal{Q}^2 $ NLI, Critic\\
\bottomrule
\end{tabular}
}
\end{center}
\caption{Hypothesis tests comparing model sizes: One-Way ANOVA for normal data, Kruskal-Wallis Test otherwise, with TukeyHSD for the former and Mann-Whitney Rank test for the latter in posthoc analysis.}
\label{tab:size_hyp_table}
\vspace{-3mm}
\end{table}

\section{Related Works}
Hallucinations in natural language generation have become a focal point of research in NLP over the past few years.
Numerous surveys \cite{ji2023survey, zhang2023siren, chen2023hallucination, li2024dawn, huang2023survey} have explored the causes, benchmarks, and mitigation strategies for hallucinations. 
\citet{luo2024hallucination} reviewed various metrics for hallucination detection, but their study did not include experiments to assess how well these metrics generalize or remain robust across different tasks and datasets.
In contrast, \citet{dziri-etal-2019-evaluating} were among the first to show that textual entailment metrics correlate more closely with human assessments of faithfulness than traditional metrics. 
Following this, \citet{honovich-etal-2021-q2} introduced $\mathcal{Q}^2$, a question-answering-based metric, which also aligns with human judgments of faithfulness.
\citet{durmus-etal-2022-spurious} pointed out that many reference-free evaluation metrics in summarization and dialogue generation rely on spurious correlations, such as word overlap, perplexity, and length, which may distort the assessment of faithfulness.
More recently, \citet{godbole2025verifycautionpitfallsrelying} highlighted that various fact-verification metrics are inconsistent and frequently misjudge system-level performance. 
Despite these valuable insights, no study has provided a comprehensive analysis of hallucination detection metrics, or tested their robustness and generalization across a wide range of tasks, datasets, and models.
The closest work to this is by \citet{kang2024comparing}, who conducted a survey of metrics within a multilingual setting.
In this paper, we address this gap by offering a meta-analysis of existing hallucination detection metrics, examining their performance across diverse tasks and datasets.

\section{Extended Discussions}
\label{sec:extended_discussion}
\subsection{Most Metrics Exhibit Poor Alignment with Human Judgment}
\label{sec:hyp_1}

\begin{table}[b]
\vspace{-5mm}
\begin{center}
\setlength{\tabcolsep}{3pt} 
\resizebox{\columnwidth}{!}{
\begin{tabular}{cccccc}
\toprule
\textbf{Dataset} & \textbf{ROUGE-L} & \textbf{SacreBleu} & \textbf{BertScore} & \textbf{Knowledge-F1} & \textbf{$\mathcal{Q}^2$ token-F1} \\
\midrule
\textsc{Begin} CMU & $-$ & $-$ & $-$ & $0.72$ & $0.70$ \\
\textsc{Begin} TC & $-$ & $-$ & $-$ & $0.75$ & $0.74$ \\
\textsc{Begin} WoW & $-$ & $-$ & $-$ & $0.43$ & $0.53$\\
\midrule
HaluEval Dial & $0.31$ & $0.32$ & $0.31$ & $0.59$ & $0.53$\\
HaluEval QA & $0.30$ & $0.54$ & $0.31$ & $0.83$  & $0.81$\\
\bottomrule
\end{tabular}%
}
\end{center}
\vspace{-2mm}
\caption{PRAUC scores between rest of the hallucination metrics and human annotations. 
}
\label{tab:extra_prauc}
\end{table}

As mentioned in \S\ref{sec:exp_setup}, we utilize the output generations from the \textsc{Begin} and HaluEval benchmarks. 
Detailed information on how the respective authors generate these responses can be found in Appendix \ref{sec:datasets}. 
\textsc{Begin} consists solely of model-generated responses and does not include gold responses, which prevents the calculation of metrics like ROUGE-L, SacreBLEU, and most of the metrics in the UniEval suite, as they are computed against the gold responses. 
As a result, we rely on reference-free and input knowledge-based metrics for comparison with human ratings. 
Although HaluEval provides gold-standard responses, we have excluded its results from Table ~\ref{tab:all_dataset_auc} to maintain consistency with the BEGIN benchmark.
Table \ref{tab:extra_prauc} provides the results (PRAUC scores) for the remaining metrics. 
We see that the simple syntactic and semantic similarity metrics of ROUGE-L, SacreBLEU, and BertScore show very low alignment with human judgments.
Knowledge-F1 and $\mathcal{Q}^2$ token-F1 yeild similar scores to Knowledge-BertScore and $\mathcal{Q}^2$-NLI F1 score.

\begin{figure}[t]
    \centering
    \includegraphics[width=\columnwidth]{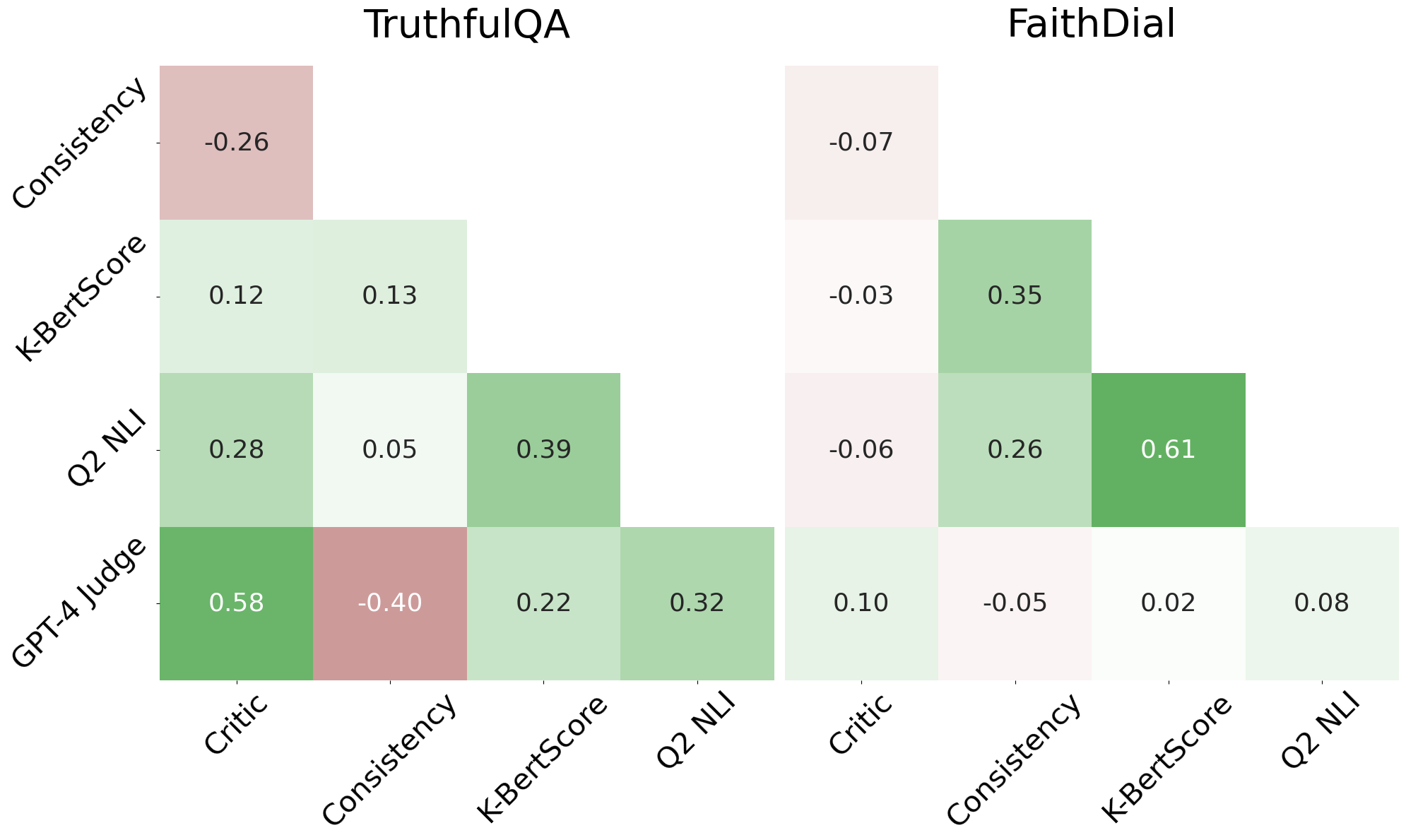}
    \caption{Spearman rank correlation between hallucination metrics reveals weak to no correlation for both TruthfulQA and FaithDial.}
    \label{fig:truthfulqa_faithdial_corr}
\end{figure}

\begin{figure}[t]
    \centering
    \includegraphics[width=\columnwidth]{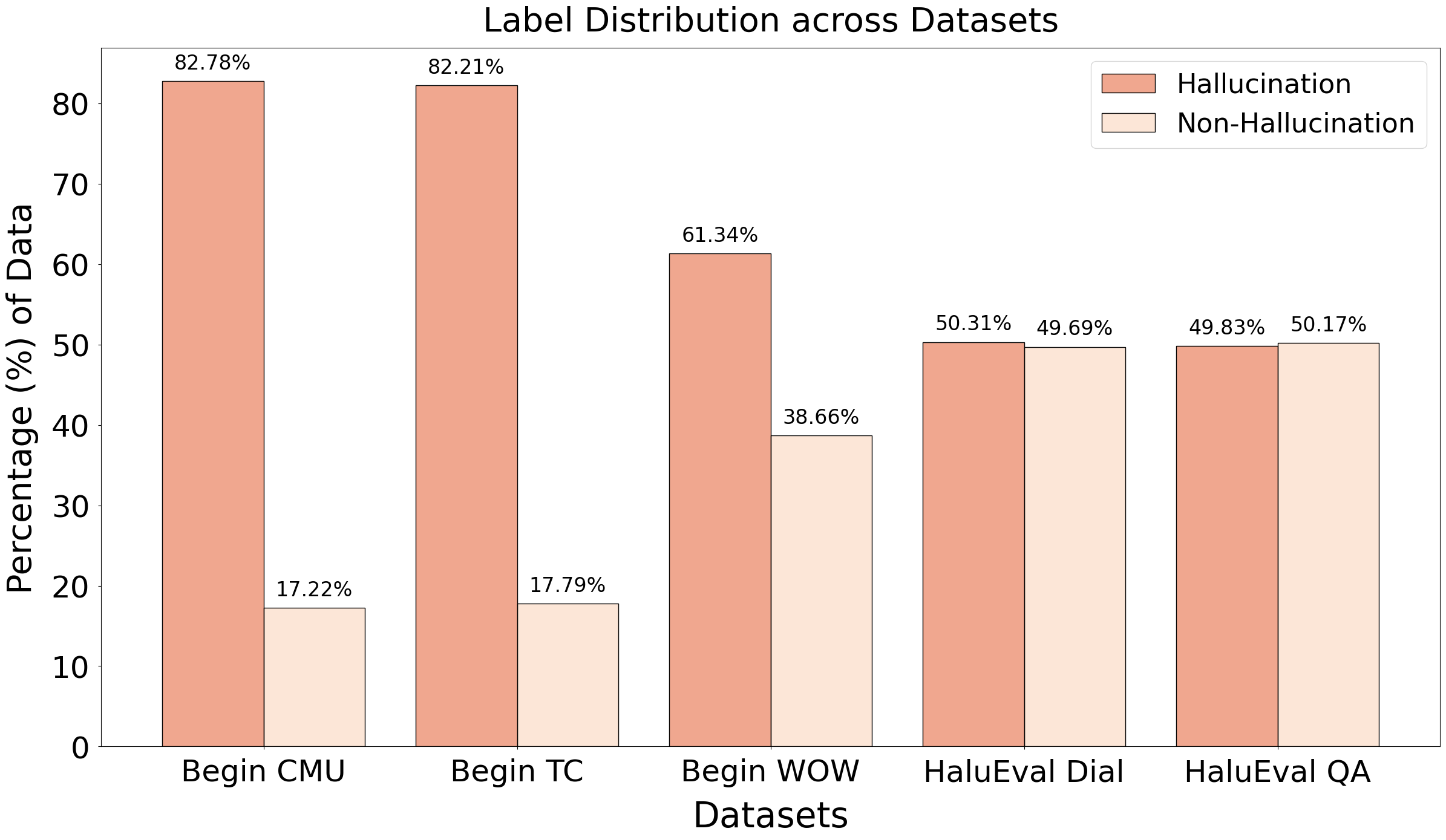}
    \caption{Distribution of hallucinated and non-hallucinated responses in \textsc{Begin} and HaluEvl.}
    \label{fig:begin_halueval_label_dist}
\end{figure}

\begin{table*}[t]
\centering
\resizebox{\textwidth}{!}{%
\begin{tabular}{ccccccccccccccccccc}
\toprule
\multirow{2}{*}{\textbf{Dataset}} & \multicolumn{3}{c}{\textbf{Critic}} & \multicolumn{3}{c}{\textbf{GPT-4}} & \multicolumn{3}{c}{\textbf{Consistency}} & \multicolumn{3}{c}{\textbf{K-BertScore}} & \multicolumn{3}{c}{\textbf{$\mathcal{Q}^2$ NLI}} & \multicolumn{3}{c}{\textbf{Ensemble}} \\
\cmidrule(lr){2-4} 
\cmidrule(lr){5-7} 
\cmidrule(lr){8-10} 
\cmidrule(lr){11-13} 
\cmidrule(lr){14-16}
\cmidrule(lr){17-19} 
 & \textbf{P} & \textbf{R} & \textbf{F1} & \textbf{P} & \textbf{R} & \textbf{F1} & \textbf{P} & \textbf{R} & \textbf{F1} & \textbf{P} & \textbf{R} & \textbf{F1} & \textbf{P} & \textbf{R} & \textbf{F1} & \textbf{P} & \textbf{R} & \textbf{F1} \\
\midrule
Begin CMU & \textcolor{Tan}{\textbf{0.77}} & 0.82 & \textcolor{Tan}{\textbf{0.77}} & \textcolor{ForestGreen}{\underline{\textbf{0.87}}} & \textcolor{ForestGreen}{\underline{\textbf{0.83}}} & \textcolor{ForestGreen}{\underline{\textbf{0.84}}} & 0.69 & 0.83 & 0.75 & 0.74 & 0.83 & 0.75 & 0.63 & 0.34 & 0.40 & 0.69 & \textcolor{Tan}{\textbf{0.83}} & 0.75 \\
Begin TC & \textcolor{Tan}{\textbf{0.70}} & 0.80 & \textcolor{Tan}{\textbf{0.74}} & \textcolor{ForestGreen}{\underline{\textbf{0.86}}} & 0.67 & 0.71 & 0.68 & 0.82 & 0.74 & 0.68 & \textcolor{Tan}{\textbf{0.82}} & 0.74 & 0.63 & 0.45 & 0.51 & 0.68 & \textcolor{ForestGreen}{\underline{\textbf{0.82}}} & \textcolor{ForestGreen}{\underline{\textbf{0.74}}} \\
Begin WoW & \textcolor{Tan}{\textbf{0.84}} & \textcolor{Tan}{\textbf{0.83}} & \textcolor{Tan}{\textbf{0.83}} & 0.81 & 0.77 & 0.77 & 0.47 & 0.61 & 0.47 & 0.38 & 0.61 & 0.47 & 0.44 & 0.46 & 0.45 & \textcolor{ForestGreen}{\underline{\textbf{0.85}}} & \textcolor{ForestGreen}{\underline{\textbf{0.85}}} & \textcolor{ForestGreen}{\underline{\textbf{0.85}}} \\
\midrule
HaluEval Dial & 0.63 & 0.56 & 0.49 & \textcolor{ForestGreen}{\underline{\textbf{0.77}}} & \textcolor{ForestGreen}{\underline{\textbf{0.74}}} & \textcolor{ForestGreen}{\underline{\textbf{0.74}}} & 0.40 & 0.40 & 0.40 & 0.60 & 0.60 & 0.60 & \textcolor{Tan}{\textbf{0.67}} & \textcolor{Tan}{\textbf{0.65}} & \textcolor{Tan}{\textbf{0.63}} & 0.46 & 0.46 & 0.45 \\
HaluEval QA & 0.54 & 0.54 & 0.53 & 0.67 & 0.66 & 0.66 & 0.43 & 0.49 & 0.36 & 0.76 & 0.76 & 0.76 & \textcolor{Tan}{\textbf{0.87}} & \textcolor{Tan}{\textbf{0.87}} & \textcolor{Tan}{\textbf{0.87}} & \textcolor{ForestGreen}{\underline{\textbf{0.87}}} & \textcolor{ForestGreen}{\underline{\textbf{0.87}}} & \textcolor{ForestGreen}{\underline{\textbf{0.87}}} \\
\midrule
Average & 0.70 & 0.71 & 0.67 & \textcolor{ForestGreen}{\underline{\textbf{0.8}}} & \textcolor{Tan}{\textbf{0.73}} & \textcolor{ForestGreen}{\underline{\textbf{0.74}}} & 0.53 & 0.63 & 0.54 & 0.63 & 0.72 & 0.66 & 0.65 & 0.55 & 0.57 & \textcolor{Tan}{\textbf{0.71}} & \textcolor{ForestGreen}{\underline{\textbf{0.77}}} & \textcolor{Tan}{\textbf{0.73}}\\
\bottomrule
\end{tabular}%
}
\caption{Weighted Precision, Recall, and F1 scores for different metrics on \textsc{Begin} and HaluEval for hallucination detection. \textcolor{ForestGreen}{\underline{Green}} and \textcolor{Tan}{brown} denote the best and second-best metrics, respectively.}
\label{tab:begin_halueval_all_results}
\end{table*}

Table \ref{tab:begin_halueval_all_results} shows the detailed classification performance of various metrics for hallucination detection on the \textsc{Begin} and HaluEval datasets.
For the \textsc{Begin} corpus, GPT-4 and the ensemble metric lead in precision, recall, and F1 scores, with Critic closely following in second place. 
However, Critic performs poorly on the HaluEval datasets.
Unsurprisingly, Critic also performs pretty well, coming in as a close second.
However, Critic performs poorly on the HaluEval datasets.
$\mathcal{Q}^2$ NLI struggles to generalize across datasets, with good performance on HaluEval, but below random chance on \textsc{Begin}, making it the second worst metric.
This contrasts with the PRAUC results in Table \ref{tab:all_dataset_auc}, where it ranks just behind Critic and the ensemble method.
UniEval’s pretrained consistency evaluator shows strong performance on \textsc{Begin} CMU and TC, but upon examining the predicted and gold label distribution in Table \ref{tab:begin_halueval_metric_label_dist} and Figure \ref{fig:begin_halueval_label_dist}, we see that the high scores are as a result its aggressive proclivity to classify everything as hallucinated.
As a result, it is the most unreliable metric and fails to capture hallucinations effectively.
K-BertScore performs poorly on \textsc{Begin} WoW and HaluEval Dial, consistent with the results in Table \ref{tab:all_dataset_auc}.

\begin{table}[t]
\centering
\resizebox{\columnwidth}{!}{%
\begin{tabular}{ccccccc}
\toprule
\textbf{Dataset} & \textbf{Critic} & \textbf{GPT4} & \textbf{Consistency} & \textbf{K-BertScore} & \textbf{$\mathcal{Q}^2$ NLI} & \textbf{Ensemble}\\
 \midrule
\textsc{Begin} CMU & 2843 / 107 & 2159 / 791 & 2949 / 1 & 22947 / 3 & 1066 / 1884 & 2949 / 1 \\
\textsc{Begin} TC & 3704 / 101 & 1993 / 1812  & 3804 / 1 & 3804 / 1 & 2069 / 1736 & 3804 / 1\\
\textsc{Begin} WoW & 1957 / 1644 & 1723 / 1878 & 3589 / 12 & 3600/1 & 2423 / 1178 & 2181 / 1420\\
\midrule
HaluEval Dial & 8686 / 1314 & 6635 / 3365 & 5492 / 4508 & 5572 / 4428 & 6863 / 3137 & 6352 / 3648 \\
HaluEval QA & 4076 / 5924 & 3712 / 6288 & 619 / 9381 & 5125/4875 & 4680/5320 & 4991/5009 \\
\bottomrule
\end{tabular}%
}
\caption{Hallucination detection label distribution (Positive/Negative) for different metrics.}
\label{tab:begin_halueval_metric_label_dist}
\vspace{-3mm}
\end{table}

\subsection{Why is the Inter-Metric Correlation Weak?}
Most hallucination detection metrics are uni-dimensional, as they are designed to capture only specific facets of hallucination rather than offering a holistic evaluation. 
This design limitation leads to low inter-metric correlation, as different metrics often emphasize fundamentally different properties of hallucinated content. 
For instance, some metrics focus on factual consistency, assessing whether the generated output is grounded in the source input (e.g., question, context, or prompt). 
Others may concentrate on fluency, semantic similarity, or entity-level accuracy.
Because these properties are orthogonal, a model might score well on one metric while performing poorly on another.

GPT-4 based evaluation is a better metric for detecting hallucinations because unlike automated metrics that rely on predefined heuristics (e.g., n-gram overlap, embeddings, or NLI classifiers), GPT-4 can assess nuanced errors, infer missing knowledge, and detect inconsistencies in a way that aligns closely with human judgment, as it considers various factors such as coherence, commonsense reasoning, and factual grounding, to name a few \citep{achiam2023gpt}. 
Here is why it shows a weak correlation with different metrics:
\begin{itemize}[leftmargin=*]
    \item \textbf{GPT-4 vs. N-gram Overlap (Rouge-L, SacreBLEU, and Knowledge-F1):} GPT-4 assesses meaning and factuality beyond simple word overlap, whereas these metrics only measure surface-level similarity. A hallucinated response can have a high n-gram overlap with a reference while still being incorrect, leading to false positives. Conversely, correct but reworded responses can be penalized, leading to false negatives. GPT-4's reasoning capabilities makes it more flexible than rigid n-gram matching.
    \item \textbf{GPT-4 vs. Semantic Similarity (BERTScore and K-BERTScore):} These metrics measure embedding similarity but do not verify factual accuracy. Two sentences can be semantically close while differing in factual correctness. GPT-4 can assess fine-grained factual inconsistencies that semantic similarity models miss, such as incorrect numerical values or subtly misleading statements.
    \item \textbf{GPT-4 vs UniEval Suite:} UniEval is trained on specific datasets and follows fixed evaluation heuristics, making it less adaptable to unseen contexts. GPT-4 dynamically evaluates responses using broad-world knowledge and deeper reasoning, leading to higher accuracy in detecting nuanced hallucinations.
    \item \textbf{GPT-4 vs. $\mathcal{Q}^2$:} It relies on question generation and answer extraction, which introduces cascading errors if the generated questions are poorly framed or if the extraction mechanism fails. Moreover, it may overlook implicit hallucinations that do not map neatly to question-answer pairs, whereas GPT-4 can reason about implicit information.
    \item \textbf{GPT-4 vs NLI-based metrics (Critic):} Critic uses a pre-trained classifier on dialogue data, meaning it lacks generalization to different domains or complex factual inconsistencies. NLI models often misinterpret negations, indirect claims, and paraphrased statements, leading to misclassifications that GPT-4 would avoid.
\end{itemize}

\subsection{Mode-Seeking Decoding Hallucinate less than Sampling-based Approaches}
\label{sec:extended_hyp_3}

\begin{figure*}[!ht]
    \centering
    \includegraphics[width=\textwidth, height=7cm]{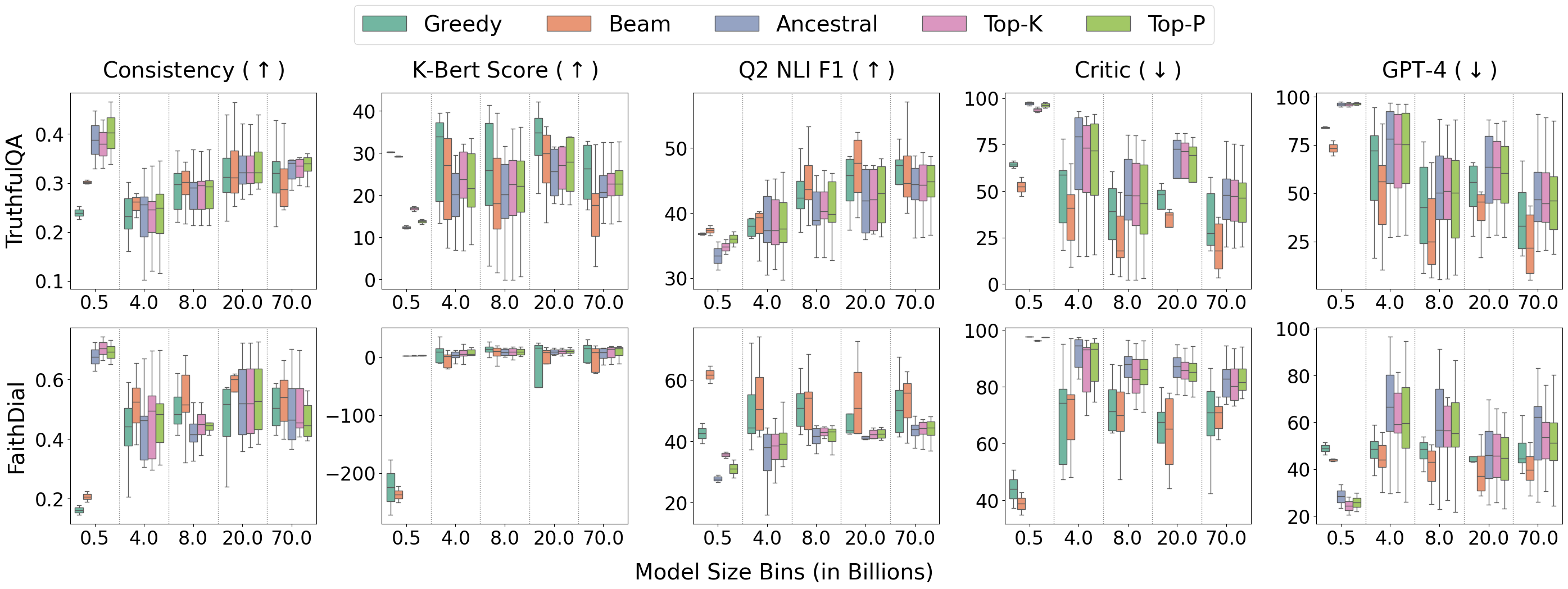}
    \caption{Comparison of factual hallucination metrics across decoding techniques. 
    }
    \label{fig:decoding_fact}
\end{figure*}

\begin{figure*}[!ht]
    \centering
    \includegraphics[width=\textwidth, height=7cm]{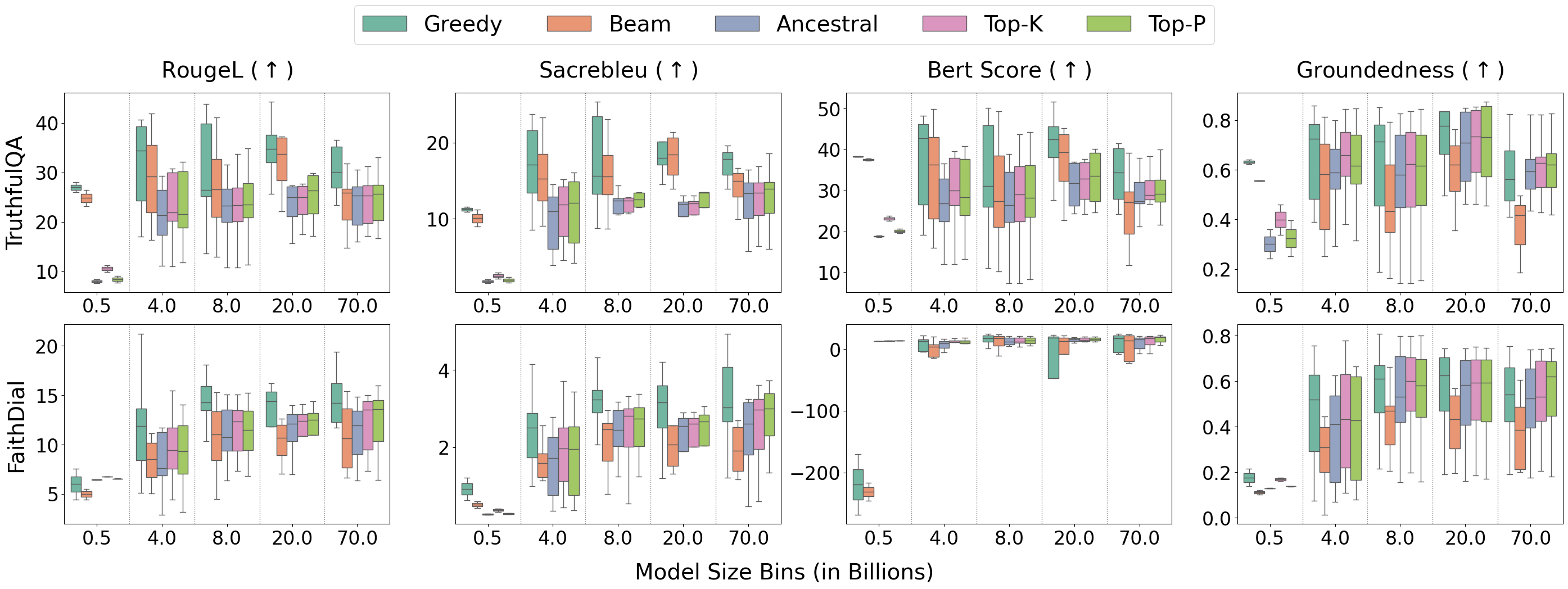}
    \caption{Comparison of traditional NLG metrics across decoding techniques.}
    \label{fig:decoding_nlg}
\end{figure*}

\begin{figure*}[!ht]
    \centering
    \includegraphics[width=\textwidth, height=7cm]{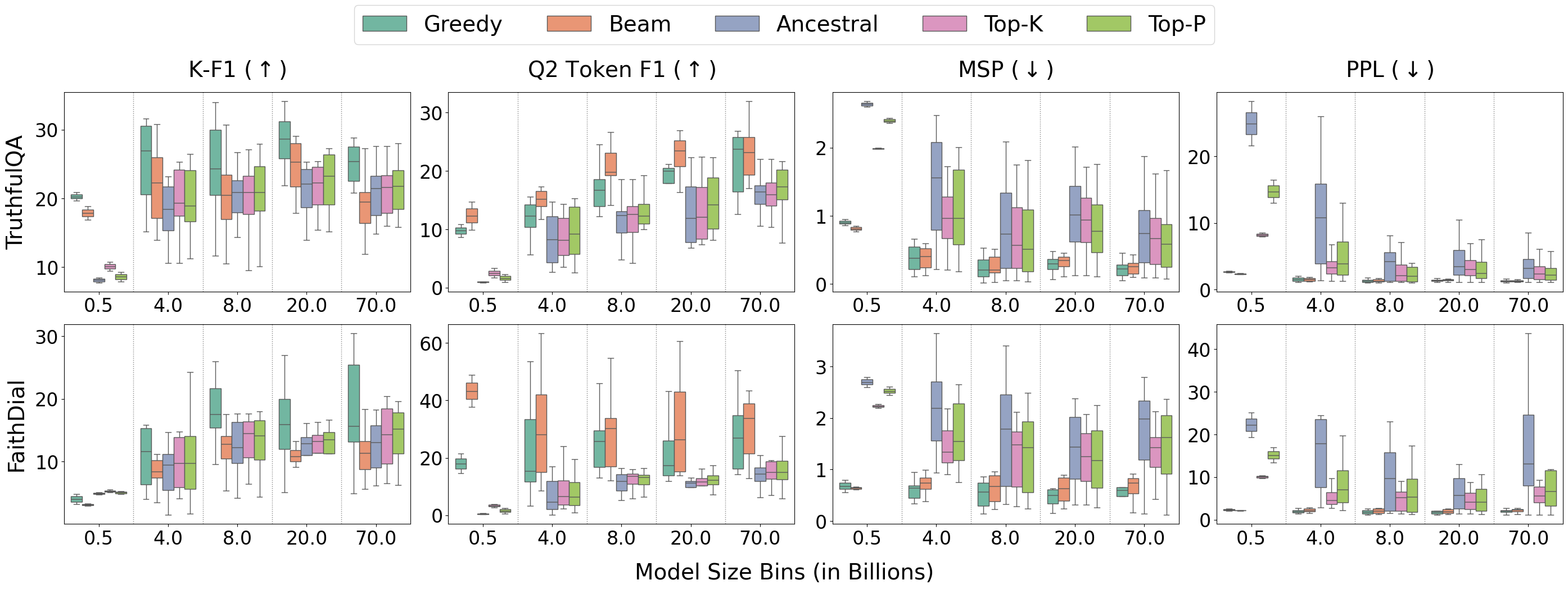}
    \caption{Comparison of uncertainty and token-overlap based hallucination metrics across decoding techniques.}
    \label{fig:decoding_other}
\end{figure*}

The box plots in Figures \ref{fig:decoding_fact}, \ref{fig:decoding_nlg}, and \ref{fig:decoding_other} illustrate the performance of various decoding techniques across different metrics.
The decoding methods considered include greedy, beam search ($b=3$), ancestral, top-k ($k=40$), and top-p ($p=0.95$). 
Models are grouped by parameter size into the following bins: ${ >0.5, >4, >20, >70}$ billion parameters.
Overall, greedy and beam search consistently outperform sampling-based methods. 
However, this trend breaks for BertScore and K-BertScore in the case of FaithDial. 
We hypothesize that this is possibly due to the model's limited capacity, which may lead to repetitive or degenerate outputs, as observed in previous studies \cite{Holtzman2020The}. 
Other metrics such as Knowledge-F1, $\mathcal{Q}^2$ token F1, MSP, and Perplexity adhere to the trend.

\begin{figure*}[!ht]
    \centering
    \includegraphics[width=\textwidth, height=9.5cm]{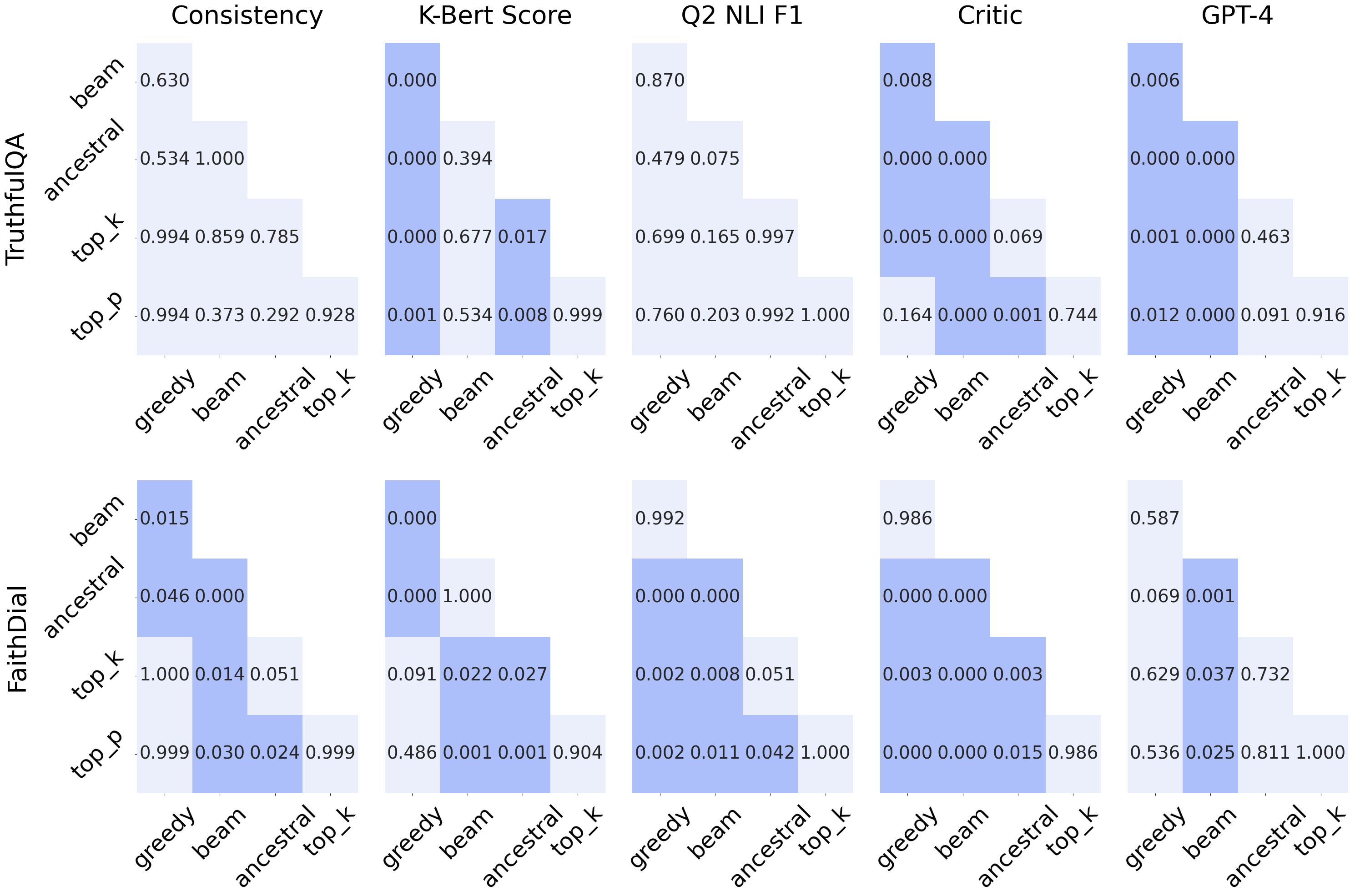}
    \caption{Per-group p-values for decoding techniques using pairwise T-test with Bonferroni correction.}   
    \label{fig:decoding_hyp_testing_fact}
\end{figure*}

\begin{figure*}[!ht]
    \includegraphics[width=\textwidth, height=9.5cm]{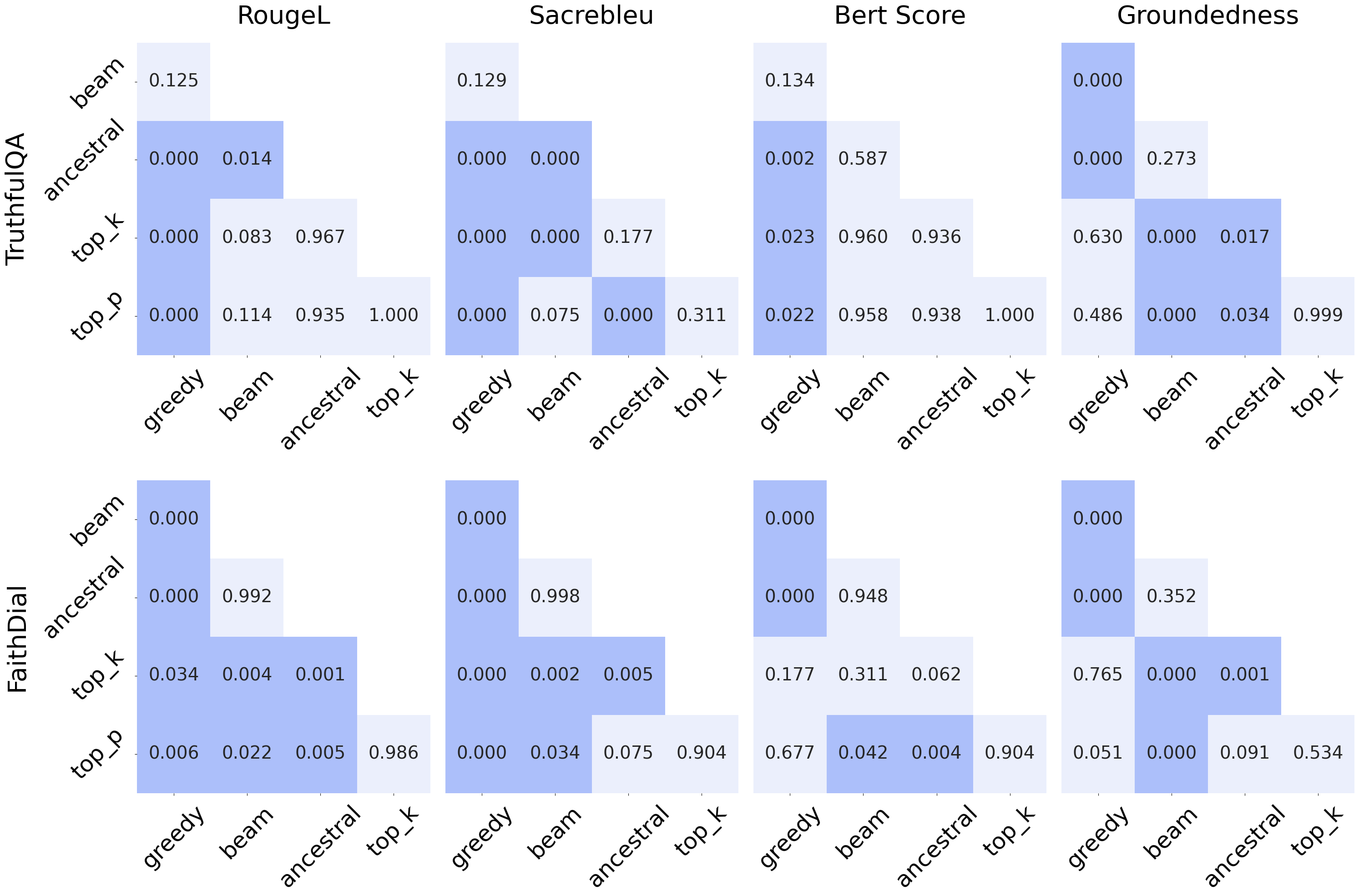}
    \caption{Per-group p-values for different decoding techniques using pairwise T-test with Bonferroni correction. }
    \label{fig:decoding_hyp_testing_nlg}
\end{figure*}

The heatmaps in Figures \ref{fig:decoding_hyp_testing_fact} and \ref{fig:decoding_hyp_testing_nlg} show the p-values for pairwise significance tests between the decoding methods. 
Except for the consistency score, greedy and beam search consistently outperform sampling-based methods with statistically significant results. 
These findings further confirm that probability-maximization decoding methods help reduce hallucinations, particularly in knowledge-grounded tasks.

\subsection{Parameter Scaling does not Necessarily Improve Hallucination Metrics}
\label{sec:extended_hyp_4}
\begin{figure*}[!ht]
    \centering
    \includegraphics[width=\textwidth, height=6cm]{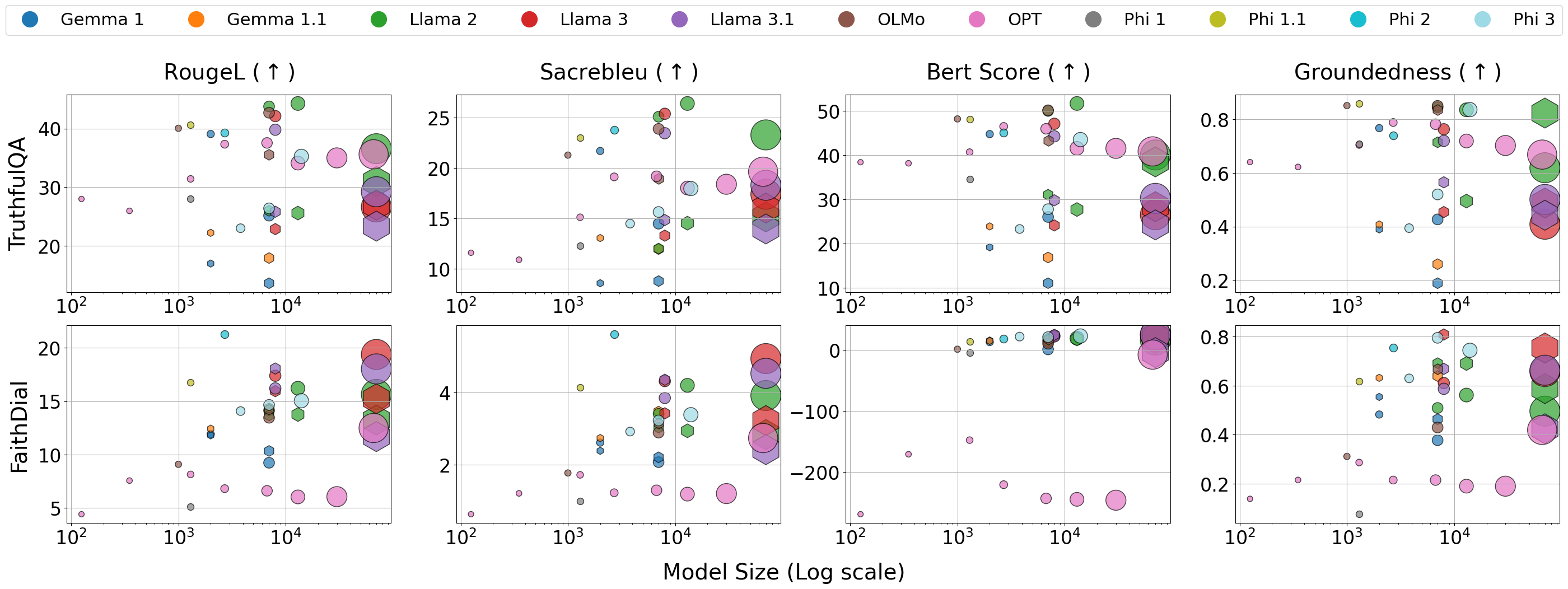}
    \caption{NLG-based hallucination detection metric scores for greedy decoding as model size increases. Circles and hexagons represent pretrained and instruction-tuned models, respectively.}
    \label{fig:scale_nlg}
\end{figure*}

\begin{figure*}[!ht]
    \centering
    \includegraphics[width=\textwidth, height=6cm]{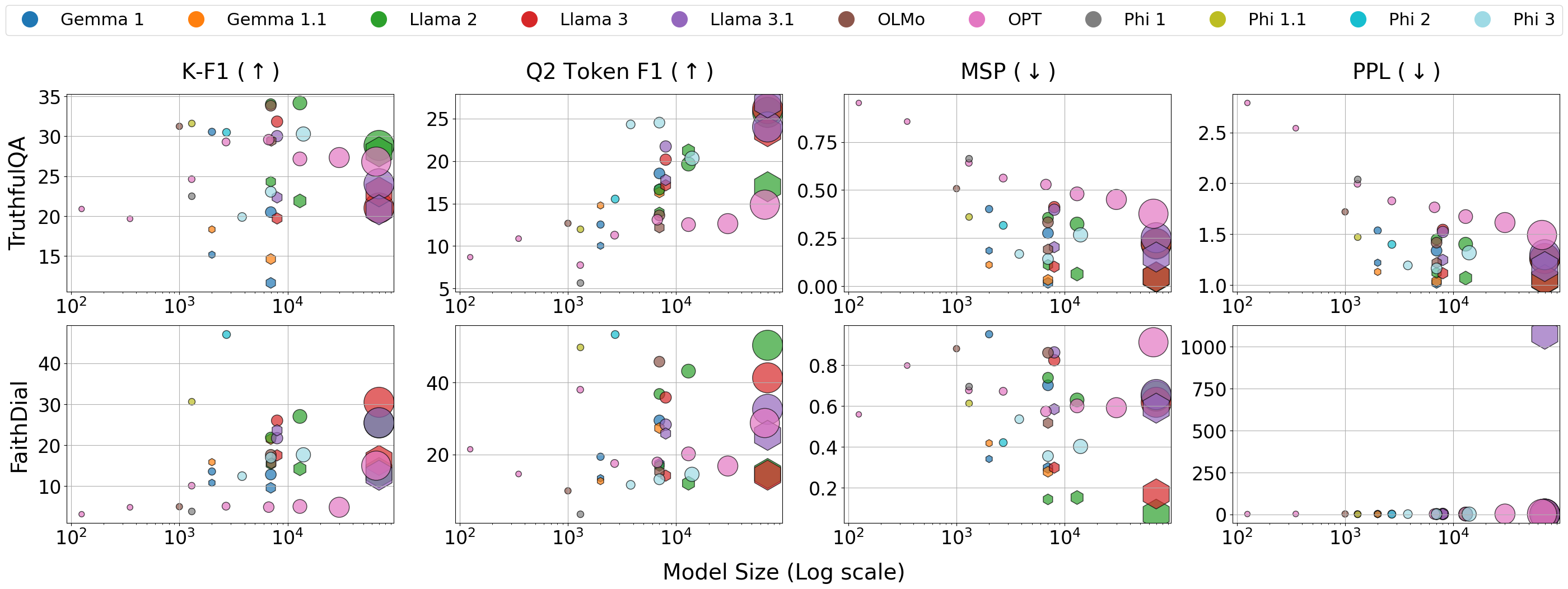}
    \caption{Uncertainty and token-overlap based hallucination detection metric scores for greedy decoding as model size increases. Circles and hexagons represent pretrained and instruction-tuned models, respectively.}
    \label{fig:scale_other}
\end{figure*}

\begin{figure*}[!ht]
    \includegraphics[width=\textwidth]{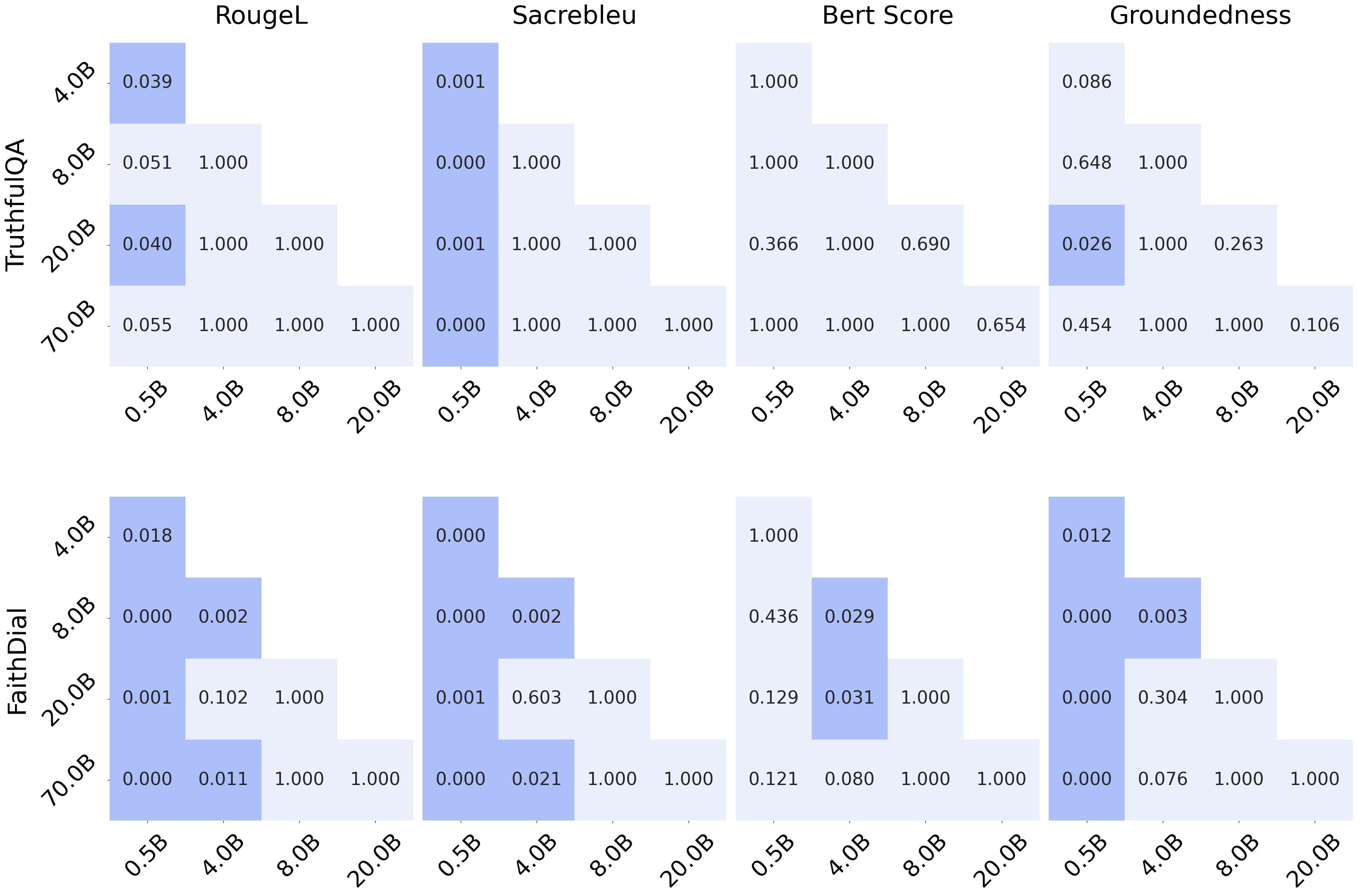}
    \caption{Per-group p-values for different model size bins using the pairwise Mann-Whitney rank test.}
    \label{fig:size_hyp_testing_nlg}
\end{figure*}

Figures \ref{fig:scale_nlg} and \ref{fig:scale_other} illustrate the performance of NLG, token-overlap, and uncertainty-based metrics as language model parameters scale.
While Rouge-L, $\mathcal{Q}^2$ token F1, MSP, and perplexity all improve with model size, other metrics do not show a consistent pattern. 
Figure \ref{fig:size_hyp_testing_nlg} presents the p-values for pairwise significance tests across different model sizes, revealing that BertScore shows no improvement as the model size increases.

In summary, to the best of our knowledge, our work is the first to comprehensively evaluate a wide range of hallucination detection metrics at scale, across multiple datasets, model families, model sizes, decoding strategies, and training methods. 
While Finding 3 have been established in prior studies, such as \citet{dziri-etal-2021-neural} and \citet{ouyang2022training}, they lack the robustness provided by our analysis, as they were not tested across the diverse dimensions that we explore. 
Our work offers a more thorough and holistic assessment, demonstrating that these findings indeed hold true across different settings and providing deeper insights for ML and NLP practitioners about which metrics perform best under various conditions.
Moreover, to the best of our knowledge, none of the previous works have concretely shown the emergence of finding 4. 
Lastly, while some of these findings might seem obvious at first, we believe scientific research is often exactly around such contributions - transforming intuitive observations into a robust, evidence-backed understanding, advancing the field with concrete, reproducible findings.

\end{document}